\documentclass[sigconf]{acmart}

\AtBeginDocument{%
  \providecommand\BibTeX{{\normalfont B\kern-0.5em{\scshape i\kern-0.25em b}\kern-0.8em\TeX}}}

\copyrightyear{2026}
\acmYear{2026}
\setcopyright{cc}
\setcctype{by}
\acmConference[MM '26]{Proceedings of the 34th ACM International Conference on Multimedia}{November 10--14, 2026}{Rio de Janeiro, Brazil}
\acmBooktitle{Proceedings of the 34th ACM International Conference on Multimedia (MM '26), November 10--14, 2026, Rio de Janeiro, Brazil}
\acmISBN{979-8-4007-2213-4/2026/11}

\usepackage{amssymb}
\usepackage{xspace}
\usepackage{multirow}
\usepackage{subcaption}
\usepackage{enumitem}
\usepackage{siunitx}

\newcommand{\method}{MSGR\xspace}
\newcommand{\methodfull}{\textbf{M}ulti-\textbf{S}cale \textbf{G}ene \textbf{R}efiner\xspace}
\newcommand{\stflowms}{STFlow+MS\xspace}
\newcommand{\egnms}{EGN+MS\xspace}

\begin{document}

\title{Gene Ontology-Guided Hierarchical Spatial Gene Expression Prediction from Histopathology Images}

\author{Zhiwen Xu}
\affiliation{%
  \institution{National University of Defense Technology}
  \city{Changsha}
  \country{China}
  }
\email{xuzhiwen@nudt.edu.cn}

\author{Xiaoming Yan}
\affiliation{%
  \institution{National University of Defense Technology}
  \city{Changsha}
  \country{China}
}
\email{yanxiaomingnudt@nudt.edu.cn}

\author{Chengkun Wu}
\affiliation{%
  \institution{National University of Defense Technology}
  \city{Changsha}
  \country{China}
  }
\email{chengkun\_wu@nudt.edu.cn}

\author{Juan Chen}
\affiliation{%
  \institution{National University of Defense Technology}
  \city{Changsha}
  \country{China}
  }
\email{juanchen@nudt.edu.cn}

\author{Haoang Chi}
\correspondingauthor
\affiliation{%
  \institution{National University of Defense Technology}
  \city{Changsha}
  \country{China}
  }
\email{haoangchi618@gmail.com}

\author{Liyang Xu}
\correspondingauthor
\affiliation{%
  \institution{National University of Defense Technology}
  \city{Changsha}
  \country{China}
  }
\email{xuliyang08@nudt.edu.cn}

\renewcommand{\shortauthors}{Zhiwen Xu et al.}

\begin{abstract}
Predicting spatial gene expression from histopathology images enables large-scale transcriptomic profiling without the cost of direct measurement. Existing methods decode the target gene set as a flat, unstructured vector, ignoring the inter-gene dependencies arising from shared biological pathways and regulatory programs. Without explicit structural guidance, models must infer these dependencies entirely from limited paired data, constraining prediction quality.
We propose \textbf{MSGR} (\textbf{M}ulti-\textbf{S}cale \textbf{G}ene \textbf{R}efiner), which bridges this gap by incorporating the Gene Ontology (GO), a curated functional hierarchy of genes, as an explicit structural prior. MSGR organizes target genes into a four-level GO tree. Its GO-guided decoder then progressively refines predictions from coarse functional domains to fine individual genes via residual corrections under scale-weighted supervision. 
Operating solely on the gene side, the GO-guided decoder serves as a seamless plug-in replacement that consistently improves existing architectures without requiring any image-side modifications.
Extensive experiments on nine datasets from the HEST-1k benchmark provide empirical evidence for two central claims: GO-structured decoding consistently outperforms flat decoding, even against a state-of-the-art generative baseline, and the gain is attributable to biological ontology structure rather than hierarchical decomposition per se, as confirmed by a +0.027 margin over a structurally equivalent random hierarchy. The source code is publicly available at \url{https://github.com/NozomiMizore/MSGR}.
\end{abstract}



\maketitle

\section{Introduction}
\label{sec:intro}

Spatial transcriptomics (ST) has emerged as a technology that enables simultaneous profiling of gene expression while preserving the spatial coordinates of measurements within tissue samples~\cite{li2021bulk}. This capability provides valuable insights into the spatial organization of cells, facilitating the characterization of cellular heterogeneity and the tumor microenvironment. Moreover, ST has demonstrated clinical relevance in elucidating disease mechanisms, discovering drug targets, and predicting therapeutic responses~\cite{lewis2021spatial,cao2024spatial,williams2022introduction}.

Although ST technologies, such as Visium~\cite{staahl2016visualization}, provide spatially resolved molecular landscapes, their clinical adoption is hindered by prohibitive costs, labor-intensive workflows, and limited sample throughput~\cite{wang2025benchmarking}. In contrast, hematoxylin and eosin (H\&E) staining---the standard protocol for preparing tissue sections for microscopic examination---is widely accessible, cost-effective, and already applied to archival collections in hospitals worldwide. These stained tissue sections are routinely digitized as whole slide images (WSIs) for diagnostic purposes. Developing a computational strategy that bridges morphological and transcriptomic modalities by predicting spatial gene expression from routinely available histopathology images could therefore transform existing WSI archives into a scalable source of molecular-level insights, reducing reliance on costly direct transcriptomic measurements~\cite{wang2025benchmarking,chelebian2025combining}.

Several computational approaches have been proposed for this cross-modal prediction task. Single-magnification regression methods extract histological features from spot-centered image patches at a single magnification and directly regress gene expression~\cite{he2020integrating,pang2021leveraging,yang2023exemplar,zeng2022spatial}. Extending this paradigm, multi-magnification regression methods enrich the histological signal by extracting features at multiple magnifications, jointly leveraging fine-grained cellular morphology and broader tissue context~\cite{chung2024accurate,wang2025m2ost,nguyen2025mmap}. Retrieval-based methods learn joint image--expression embeddings via contrastive learning and impute predictions from nearest neighbors during inference~\cite{xie2023spatially}. More recently, generative modeling approaches have formulated this problem as a conditional generation task~\cite{huang2025scalable,zhu2025diffusion}.

\begin{figure}[t!]
\centering
\includegraphics[width=\linewidth]{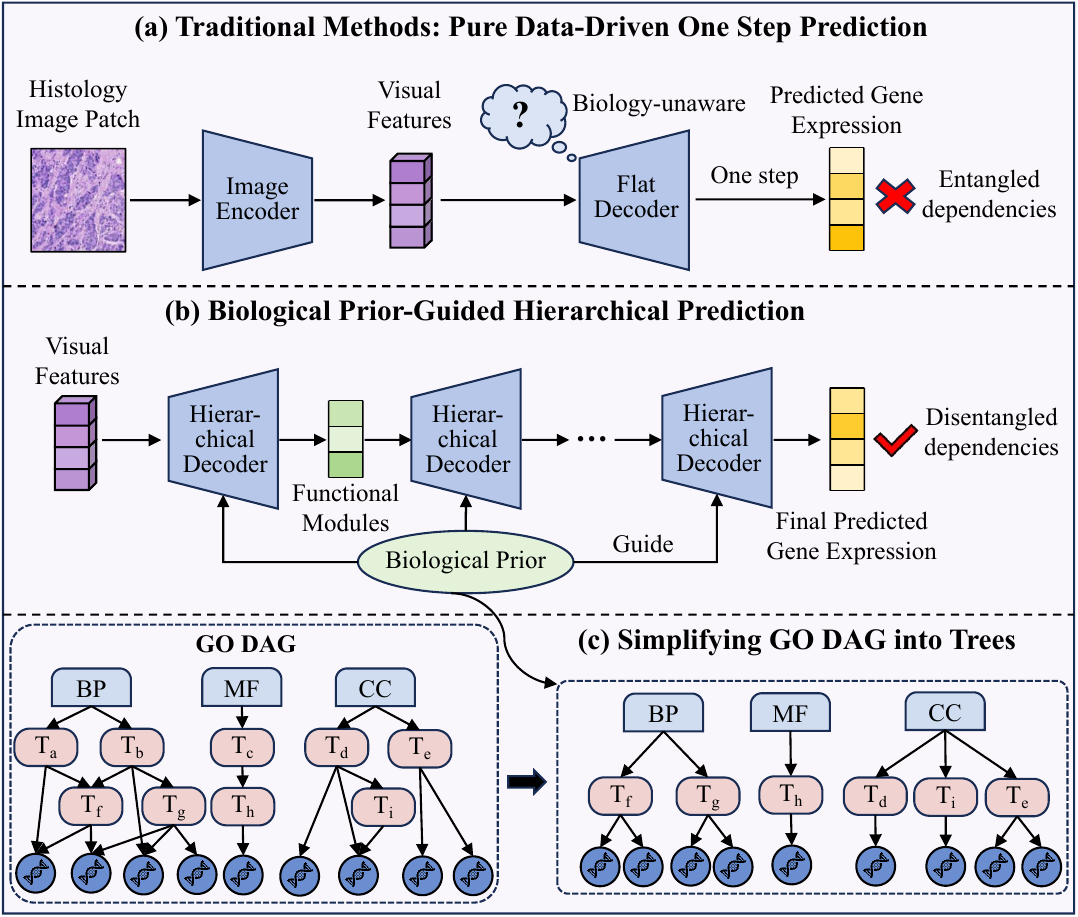}
\caption{%
    Traditional flat decoding vs.\ hierarchical decoding guided by a biological prior.
    (a) Flat decoder outputs all genes simultaneously as a flat vector.
    (b) Biological prior-guided decoder refines predictions coarse-to-fine.
    (c) Simplifying the complex GO DAG architecture into tractable trees.
}
\Description{Comparison of flat gene decoding with GO-guided coarse-to-fine decoding, alongside a diagram showing how the GO directed acyclic graph is simplified into trees.}
\label{fig:intro}
\end{figure}

Despite their architectural diversity, existing methods share a fundamental limitation: they decode all target genes simultaneously, reducing an inherently structured gene set to a flat output vector (Figure~\ref{fig:intro}(a)). Such a design does not explicitly account for the cross-modal semantic gap between tissue-level morphological phenotypes captured by histopathology images and fine-grained molecular variation reflected in gene expression. It also requires the model to implicitly learn complex functional and expression dependencies among hundreds of genes from limited paired data. Introducing structured biological knowledge as an intermediate representation offers a natural alternative, transforming flat, one-step decoding into a coarse-to-fine prediction process that progressively narrows the semantic gap while explicitly coordinating the predictions of functionally related genes (Figure~\ref{fig:intro}(b)).

The Gene Ontology (GO)~\cite{ashburner2000gene} provides a structured representation of biological knowledge. As a curated vocabulary, GO annotates genes through a hierarchical framework modeled as a directed acyclic graph (DAG). This framework partitions biological concepts into three top-level domains (biological process, molecular function, and cellular component), which branch into progressively more specific descendant terms. Importantly, genes sharing granular GO annotations tend to exhibit correlated expression patterns~\cite{van2018gene}, providing a biological foundation for structured prediction. While the native GO architecture is a complex DAG (Figure~\ref{fig:intro}(c)), it can be projected into a computationally tractable tree structure~\cite{liang2018hidden,gu2024simona}, enabling a unique ancestral path for hierarchical decoding. Unlike protein-protein interaction (PPI) networks (e.g., STRING~\cite{szklarczyk2025string}), which primarily represent non-hierarchical pairwise interactions, the GO framework provides a curated, top-down hierarchy, making it naturally suited for the hierarchical decoding paradigm.

Motivated by these observations, we present \textbf{\method} (\methodfull), a multi-scale spatial gene prediction framework simplifying the GO DAG into hierarchical trees as structural priors. This approach decomposes the regression of all target genes into a sequence of biologically aligned subtasks. Instead of treating all target genes as a flat output vector, \method organizes them into a four-level GO tree (virtual root $\rightarrow$ GO domains $\rightarrow$ GO terms $\rightarrow$ genes) and progressively refines predictions from the coarsest level to the finest. At each non-root level, a shared Transformer backbone predicts a residual correction to the estimate inherited from the coarser level. Concurrently, a cross-scale latent highway propagates latent context from coarser levels to finer ones, providing each refinement step with direct access to the broader biological context that guides its correction. By decoupling structured gene decoding from the image-feature extractor, the GO-guided decoder can be integrated into different base architectures. Extensive experiments on nine HEST-1k~\cite{jaume2024hest} datasets demonstrate that \method outperforms existing methods, while integrating the decoder into existing architectures yields consistent performance improvements. The main contributions of this work are as follows:

\begin{itemize}[leftmargin=*,noitemsep,topsep=2pt]
    \item We introduce \method, a novel hierarchical framework for spatial gene expression prediction. By progressively refining predictions across a GO-structured hierarchy, \method effectively bridges the cross-modal semantic gap between tissue morphology and fine-grained molecular events, transforming conventional flat decoding into a principled, coarse-to-fine prediction process.
    \item We propose a systematic scheme to construct a biologically aligned four-level GO hierarchy applicable to arbitrary target gene sets. This method transforms curated biological knowledge into a computationally tractable prior, enabling the model to explicitly exploit functional interdependencies among genes. This design shifts the burden from purely data-driven implicit learning to knowledge-guided structured inference.
    \item Benchmarks on nine HEST-1k datasets show that \method achieves state-of-the-art performance. Importantly, the GO-guided multi-scale decoder exhibits exceptional model-agnostic versatility. As a plug-and-play enhancer, it consistently yields performance gains when integrated into various leading architectures, validating its superior capability in knowledge-guided cross-modal prediction.
\end{itemize}
\section{Related Work}
\label{sec:related}

\subsection{Spatial Gene Expression Prediction from Histopathology Images}

Early approaches typically formulate spatial gene expression prediction as a supervised multi-output regression problem based on image features. ST-Net~\cite{he2020integrating} predicts spatially resolved gene expression from convolutional neural network (CNN) features. HisToGene~\cite{pang2021leveraging} extends this paradigm with a Vision Transformer (ViT)~\cite{dosovitskiy2020image} that jointly models all spots via self-attention~\cite{vaswani2017attention}. To model tissue structures, Hist2ST~\cite{zeng2022spatial} extracts patch features using a ConvMixer~\cite{trockman2022patches} module, then leverages Transformers~\cite{vaswani2017attention} and Graph Neural Networks (GNNs)~\cite{scarselli2008graph} to capture global dependencies and local spatial interactions. THItoGene~\cite{jia2024thitogene} further combines dynamic convolution with capsule networks~\cite{sabour2017dynamic} and Graph Attention Networks (GATs)~\cite{velivckovic2017graph} to refine predictions through spot-interaction modeling. Distinct from these direct regression architectures, EGN~\cite{yang2023exemplar} introduces an ``Exemplar Learning'' mechanism within the regression framework. It refines intermediate visual features by retrieving and interacting with visually similar reference samples before feeding the enhanced features into the regression head. BLEEP~\cite{xie2023spatially} adopts a retrieval-based approach, learning a joint image--expression embedding via contrastive learning and imputing predictions from nearest neighbors during inference. ST-Align~\cite{lin2024st} similarly employs contrastive learning to align gene and histological features at both spot and niche spatial scales, using scGPT~\cite{cui2024scgpt} and UNI~\cite{chen2024towards} to extract gene and image features, respectively.

More recently, generative models have been applied to capture the stochastic nature of gene expression. Stem~\cite{zhu2025diffusion} employs a conditional diffusion model~\cite{ho2020denoising} built on a DiT~\cite{peebles2023scalable} backbone, conditioning on UNI~\cite{chen2024towards} features to learn the one-to-many mapping from histopathology to gene expression. STFlow~\cite{huang2025scalable} adopts flow matching~\cite{lipman2022flow} to model the joint distribution of gene expression across all spots within a whole slide. It further incorporates an E(2)-invariant frame averaging mechanism~\cite{puny2021frame} to facilitate efficient spatial modeling. Existing methods across these paradigms generally retain a flat gene-decoding step: no prior is imposed on the functional relationships among genes. To the best of our knowledge, \method is the first method to use GO as a structural prior for coarse-to-fine spatial gene expression decoding from histopathology images.

\subsection{Image-Side Multi-Magnification Methods}

A complementary research direction enriches the histological signal by processing patches at multiple magnifications. TRIPLEX~\cite{chung2024accurate} processes each spot at three resolutions---the spot patch, a local neighborhood, and a global tissue context patch---and fuses these representations through cross-attention. Similarly, M2OST~\cite{wang2025m2ost} employs a many-to-one Transformer architecture to jointly process patches from three magnification levels. Additionally, MMAP~\cite{nguyen2025mmap} captures fine-grained morphological details by generating high-resolution sub-patches through random cropping and fusing these multi-magnification features through cross-attention. It further incorporates global tissue context from a slide-level prototype bank. Although these methods improve predictive performance by enriching histological signals, the gene decoding process relies on a flat structure. Crucially, the proposed gene-side multi-scale hierarchy operates along a fundamentally different dimension from these image-side multi-magnification approaches: \method decomposes the gene output space using the GO structure, whereas the aforementioned methods enrich the image input space. These two paradigms are complementary and can be effectively integrated.
\section{Method}
\label{sec:method}

\subsection{Problem Formulation}

A tissue slide contains $N$ spots, each characterized by (i) a histology image patch $\mathbf{I}_i \in \mathbb{R}^{H \times W \times 3}$, where $H$ and $W$ denote the height and width of the image patch, respectively; (ii) a spatial coordinate $\mathbf{C}_i \in \mathbb{R}^2$; and (iii) a log1p-transformed expression vector $\mathbf{y}_i \in \mathbb{R}^G$, where $G$ denotes the number of target genes. The objective is to predict $\mathbf{y}_i$ from $(\mathbf{I}_i,\mathbf{C}_i,
\{\mathbf{I}_j,\mathbf{C}_j\}_{j\in\mathcal{N}(i)})$, where $\mathcal{N}(i)$ denotes the set of spatial neighbors of spot $i$, excluding spot $i$ itself.

\subsection{Overview of MSGR}
\label{sec:overview}

\begin{figure*}[h!]
    \centering
    \includegraphics[width=0.95\linewidth]{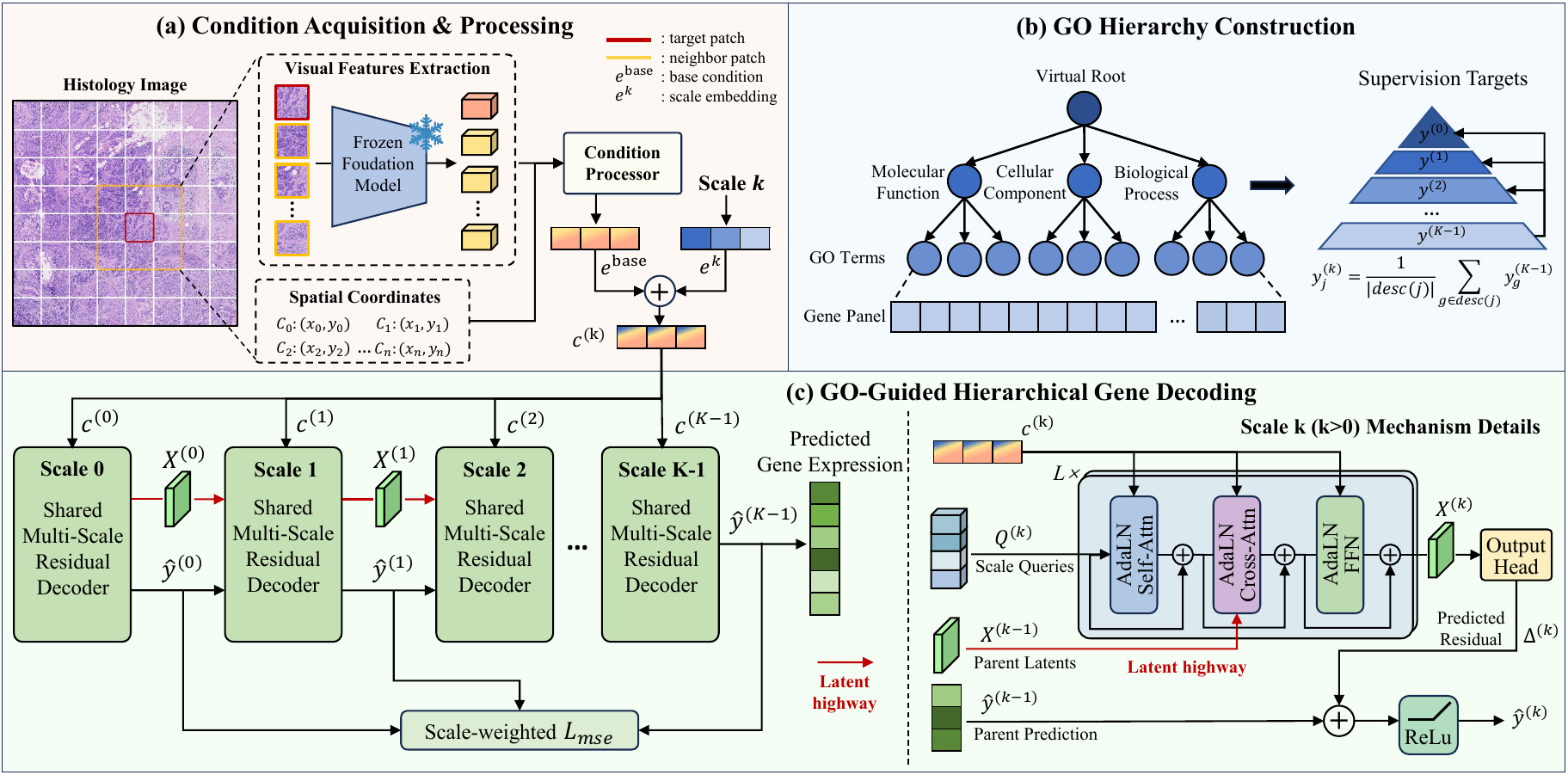}
    \caption{
        Overview of the \method framework.
        (a)~Condition Acquisition \& Processing: A frozen UNI encoder extracts per-spot patch embeddings, which are enriched with local neighborhood context and fused with spatial coordinates to form a scale-aware conditioning vector.
        (b)~GO Hierarchy Construction: Target genes are organized into a four-level tree; coarser-level supervision targets are computed as the mean expression of descendant leaf genes.
        (c)~GO-Guided Hierarchical Decoding: The decoder traverses the hierarchy from root to leaf genes. At scale 0, the cross-attention sublayer in AdaLNCrossScaleBlock attends only to the condition token, whereas at finer scales it also attends to parent-scale latent tokens through the latent highway (red line). The output head directly produces the root prediction at scale 0 and produces a residual correction at each finer scale.
    }
    \Description{Architecture diagram of MSGR: frozen UNI image features are processed with local spatial context and coordinates; target genes are organized in a four-level GO tree; and a GO-guided residual decoder refines predictions from the virtual root to individual genes using a cross-scale latent highway.}
    \label{fig:overview}
\end{figure*}

An overview of the proposed \method is presented in Figure~\ref{fig:overview}. The core idea is to decompose the prediction of hundreds of genes into a sequence of biologically aligned sub-tasks organized by the GO hierarchy, enabling the model to leverage functional relationships among genes rather than treating them as independent outputs. Histopathology patch embeddings are initially extracted by a frozen UNI encoder~\cite{chen2024towards}, a foundation model pre-trained on over 100,000 WSIs that provides robust and transferable visual representations. These embeddings are subsequently fed into the Condition Processor, which aggregates local tissue context from neighboring spots via a Spatial Transformer and fuses the enriched representations with spatial coordinates to produce a scale-aware conditioning vector. The GO hierarchy is constructed offline by mapping target genes to the most specific GO annotations and organizing them into a four-level tree, which spans from a virtual root through GO domains and GO terms to individual leaf genes. The multi-scale residual decoder then leverages this hierarchy as structural guidance, traversing the tree level by level and applying conditioning-driven residual corrections to progressively refine predictions from coarse functional domains to individual genes. Detailed descriptions of each component are provided in the subsequent sections.

\subsection{Condition Acquisition \& Processing}
\label{sec:cond}

The condition acquisition and processing pipeline begins by extracting per-spot patch embeddings using a frozen UNI encoder, as depicted in Figure~\ref{fig:overview}(a). For each spot $i$, the UNI encoder produces a histopathology embedding $\mathbf{p}_i \in \mathbb{R}^{D_h}$, which is projected to a target dimension $d$ via a linear layer. The embeddings of the center spot and its $n$ nearest spatial neighbors are then concatenated to form a token sequence of $n{+}1$ vectors in $\mathbb{R}^d$. Inspired by STFlow~\cite{huang2025scalable}, a Spatial Transformer applies $L_s$ layers of E(2)-equivariant message passing via frame averaging~\cite{puny2021frame}. Specifically, the relative coordinates of each center--neighbor pair are transformed into a canonical frame to obtain rotation-invariant geometric representations. Edge features are then constructed from the frame-projected coordinates and the inter-spot distance. A learned MLP-attention mechanism uses the edge features to compute attention weights for aggregating neighboring representations (see Appendix~B for the detailed architecture). The resulting center token output, $\mathbf{h}_i^{\text{spa}} \in \mathbb{R}^{d}$, provides a spatially aware representation that captures both the histopathological content and the local tissue context.

The spatially enriched representation $\mathbf{h}_i^{\text{spa}}$ and 2D coordinate $\mathbf{C}_i$ are projected through the histopathology and coordinate branches,
$\phi_h(\mathbf{h})=\text{SiLU}(\text{Linear}(\text{LN}(\mathbf{h})))$ and
$\phi_C(\mathbf{C})=\text{SiLU}(\text{Linear}(\mathbf{C}))$, respectively, and fused to form the base condition:
\begin{equation}
\mathbf{c}_i^{\text{base}}
= \text{Dropout}\left(
\text{LN}\left(
\phi_h(\mathbf{h}_i^{\text{spa}})
+ \phi_C(\mathbf{C}_i)
\right)\right).
\label{eq:cond}
\end{equation}
A learnable scale embedding is then added at each scale $k$:
\begin{equation}
\mathbf{c}_i^{(k)}
= \mathbf{c}_i^{\text{base}}
+ \mathbf{e}_k^{\text{scale}},
\qquad
\mathbf{e}_k^{\text{scale}} \in \mathbb{R}^{d},
\end{equation}
enabling the shared conditioning representation to encode scale-specific information without duplicating conditioning parameters.

\subsection{GO Hierarchy Construction}
\label{sec:go_hier}
Given a target gene set of $G$ genes, a four-level tree $\mathcal{T}$ is constructed using GO annotations, as illustrated in Figure~\ref{fig:overview}(b):
\begin{enumerate}[leftmargin=*,noitemsep,topsep=2pt]
    \item Level 0 (Virtual root): A single node that aggregates all $G$ genes through its descendant nodes;
    \item Level 1 (GO domains): Up to four nodes, including three GO domain roots (biological process, molecular function, and cellular component) and an optional unannotated proxy node;
    \item Level 2 (GO terms and proxy nodes): The direct children of each domain root following path compression. The total count is data-dependent (typically around 20 for $G{=}200$). Completely unannotated genes are positioned under a Level 2 proxy node associated with the Level 1 unannotated branch;
    \item Level 3 (Individual genes): $G$ leaf nodes, with each target gene assigned exactly once.
\end{enumerate}

The tree is constructed offline in four steps. (i)~Annotation query: Mappings from GO terms to genes are retrieved for the target gene set using g:Profiler~\cite{kolberg2023g}. Each gene is assigned to its most specific annotated term (i.e., the term with the smallest term size) as its representative node. (ii)~Tree projection: For each term possessing multiple parents in the GO DAG, the direct parent with the largest term size is retained. This preference for broader parent terms preserves greater semantic coverage at coarser hierarchy levels. Parent links are then traced upward to connect terms to their domain roots. (iii)~Path compression: Intermediate nodes possessing a single child and lacking directly assigned genes are merged, keeping the term layer shallow. (iv)~Proxy routing: Genes lacking a matching GO term are routed to proxy nodes, ensuring that every gene appears exactly once at Level 3. In the exported hierarchy, GO terms are flattened to a single layer beneath each domain root, while deeper descendants are aggregated into the nearest retained term or proxy node. Coarser-scale supervision targets are computed as the mean expression of descendant leaf genes:
\begin{equation}
    \mathbf{y}^{(k)}_j = \frac{1}{|\text{desc}(j)|} \sum_{g \in \text{desc}(j)} \mathbf{y}^{(K-1)}_g,
    \quad 0 \leq k < K-1,
    \label{eq:target}
\end{equation}
where $K=4$ denotes the total number of hierarchy levels, and $\text{desc}(j)$ represents the set of leaf genes descended from node $j$.

\subsection{GO-Guided Hierarchical Decoding}
\label{sec:decoder}

As illustrated in Figure~\ref{fig:overview}(c), the multi-scale residual decoder carries out the core refinement process of \method. Rather than predicting all $G$ genes simultaneously in a single step, it traverses the GO hierarchy from scale $k=0$ (virtual root) to scale $K-1$ (individual genes). At each level $k>0$, the parent-scale estimate is refined by a learned residual correction:
\begin{equation}
    \hat{\mathbf{y}}^{(k)} \;=\; \text{ReLU}\!\left(\hat{\mathbf{y}}^{(k-1)}\!\left[\text{par}^{(k)}\right] + \boldsymbol{\Delta}^{(k)}\right),
\end{equation}
where $\hat{\mathbf{y}}^{(k-1)}[\text{par}^{(k)}]$ broadcasts each parent's prediction to its children, $\boldsymbol{\Delta}^{(k)}$ denotes the residual correction learned at scale $k$, and ReLU enforces non-negativity. The complete formulation including the special case at $k=0$ is provided in Equation~\eqref{eq:residual}. This decomposition mirrors the organization of biological functions: a global root-level estimate is first obtained, then refined into GO domain-level and GO term-level activities, and ultimately into individual gene expression values.

\emph{Scale queries and shared backbone.}
The proposed framework utilizes $K{=}4$ hierarchy levels, as defined in Section~\ref{sec:go_hier}. Let $|\mathcal{S}_k|$ denote the number of nodes at scale $k$. The root level $|\mathcal{S}_0|{=}1$ is fixed, the leaf level $|\mathcal{S}_{K-1}|{=}G$ equals the number of target genes, and the intermediate levels $|\mathcal{S}_1|, |\mathcal{S}_2|$ are determined by the GO annotations of the specific target gene set. At each scale $k$, a dedicated set of learnable semantic queries $\mathbf{Q}^{(k)} \in \mathbb{R}^{|\mathcal{S}_k| \times d}$ is introduced. These queries are processed by $L$ shared AdaLNCrossScaleBlocks, where $L$ is the number of Transformer layers:
\begin{equation}
    \mathbf{X}^{(k)} = \text{DecBlock}_L \!\circ \cdots \circ\, \text{DecBlock}_1\!
    \left(\mathbf{Q}^{(k)},\; \mathbf{c}_i^{(k)},\; \text{ctx}^{(k)}\right),
    \label{eq:backbone}
\end{equation}
where sharing parameters across all scales provides cross-scale regularization and reduces the total parameter count. The decoder comprises four tightly coupled components: (i)~per-scale learnable semantic queries $\mathbf{Q}^{(k)}$ that encode the identity of nodes at each hierarchy level; (ii)~a shared AdaLNCrossScaleBlock backbone that updates these queries under histopathology conditioning; (iii)~a latent highway that supplies each scale with the context of the parent scale via cross-attention; and (iv)~a shared scalar output head that maps each updated token to a residual correction.

\emph{AdaLNCrossScaleBlock.}
Each block applies three sub-layers conditioned by Adaptive Layer Normalization (AdaLN)~\cite{peebles2023scalable}, where a single $\text{SiLU}{\to}\text{Linear}(d,9d)$ projection of $\mathbf{c}_i^{(k)}$ generates scale, shift, and gate parameters for each sub-layer:
\begin{enumerate}[leftmargin=*,noitemsep,topsep=2pt]
    \item Self-attention over the $|\mathcal{S}_k|$ hierarchy-node tokens, with $\ell_2$-normalized queries and keys for training stability;
    \item Cross-attention over the context $\text{ctx}^{(k)}$ (as detailed in the latent-highway formulation below);
    \item SwiGLU FFN: $(\mathbf{x}\mathbf{W}_1 \odot \text{SiLU}(\mathbf{x}\mathbf{W}_2))\mathbf{W}_3$.
\end{enumerate}

\emph{Latent highway with condition token injection.}
The cross-attention context for each scale is formulated as:
\begin{equation}
    \text{ctx}^{(k)} =
    \begin{cases}
        \mathbf{c}_i^{(k)} & k = 0 \\
        \left[\mathbf{X}^{(k-1)} \;\|\; \mathbf{c}_i^{(k)}\right] & k > 0
    \end{cases},
    \label{eq:ctx}
\end{equation}
where $\mathbf{c}_i^{(k)}$ serves as the condition token. At scale 0 (which lacks a parent), the condition token alone serves as the cross-attention key/value sequence. At scales $k > 0$, the parent-scale latents $\mathbf{X}^{(k-1)}$ are concatenated with the condition token, providing the current scale with direct access to both the parent-scale latent context (via the latent highway) and the histopathological condition.

\emph{Residual output and ReLU non-negativity.}
The scalar output head, shared across scales, maps each node token in $\mathbf{X}^{(k)}$ to a scalar, collectively forming $\boldsymbol{\Delta}_i^{(k)} \in \mathbb{R}^{|\mathcal{S}_k|}$. At $k=0$, the single output serves as the root prediction; at finer scales, the node-wise outputs provide residual corrections to the inherited parent estimates:
\begin{equation}
    \hat{\mathbf{y}}_i^{(k)}
    =
    \begin{cases}
        \operatorname{ReLU}\!\left(
            \boldsymbol{\Delta}_i^{(0)}
        \right),
        & k = 0, \\[2pt]
        \operatorname{ReLU}\!\left(
            \hat{\mathbf{y}}_i^{(k-1)}
            \!\left[\operatorname{par}^{(k)}\right]
            + \boldsymbol{\Delta}_i^{(k)}
        \right),
        & k > 0,
    \end{cases}
    \label{eq:residual}
\end{equation}
where $\operatorname{par}^{(k)}$ broadcasts parent predictions to their children, and ReLU enforces non-negativity. The final prediction for spot $i$ is $\hat{\mathbf{y}}_i=\hat{\mathbf{y}}_i^{(K-1)} \in \mathbb{R}^{G}$. As the decoder only replaces the final linear head, it can be integrated into diverse upstream conditioning architectures with minimal modification.

\subsection{Training Objective}

The multi-scale MSE loss with hierarchy-level supervision is formulated as:
\begin{equation}
    \mathcal{L} = \sum_{k=0}^{K-1} \bar{w}_k \cdot \mathrm{MSE}\!\left(\hat{\mathbf{y}}^{(k)},\, \mathbf{y}^{(k)}\right),
    \quad \bar{w}_k = \frac{w_k}{\sum_{k'} w_{k'}},
    \label{eq:loss}
\end{equation}
where $\bar{w}_k$ denotes the normalized weight for scale $k$. These weights allow the training objective to place greater emphasis on finer scales, particularly the final gene level, while downweighting coarser-scale supervision that may be relatively noisy (Appendix~C.1).

\section{Experiments}
\label{sec:exp}
We conduct experiments on nine datasets to address the following questions:
\textbf{RQ1:} Does \method outperform existing flat-decoding methods?
\textbf{RQ2:} Can the GO-guided decoder improve existing architectures without image-side modifications?
\textbf{RQ3:} Does \method faithfully recover the spatial expression patterns of cancer marker genes?
\textbf{RQ4:} Do the proposed architectural components contribute to performance?
\textbf{RQ5:} Are the gains attributable to the GO-derived hierarchy rather than hierarchical decomposition alone?
\textbf{RQ6:} How sensitive is \method to variations in its hyperparameters?
\subsection{Setup}

\subsubsection{Datasets.}
We evaluate \method and baseline methods on nine spatial transcriptomics datasets from HEST-1k~\cite{jaume2024hest}, i.e., CCRCC, COAD, HCC, IDC, LUNG, PRAD, READ, SKCM, and KIDNEY. The detailed dataset statistics are provided in Appendix~A.
We apply a patient-stratified split to prevent data leakage, thereby establishing a $k$-fold cross-validation setup. Each dataset contains multiple slides distributed across independent folds (2--6 folds).
For each dataset, the top 200 genes are selected from the intersection of highly expressed and highly variable genes for evaluation, following the preprocessing protocol introduced by Stem~\cite{zhu2025diffusion}.

\subsubsection{Baselines.}
Comparisons are conducted against seven published methods that span four categories:
\begin{itemize}[leftmargin=*,noitemsep,topsep=2pt]
    \item Single-magnification regression: ST-Net~\cite{he2020integrating} (CNN), EGN~\cite{yang2023exemplar} (exemplar-guided ViT), HisToGene~\cite{pang2021leveraging} (ViT + spatial PE);
    \item Retrieval-based: BLEEP~\cite{xie2023spatially} (contrastive embedding + $k$-NN imputation);
    \item Multi-magnification regression: TRIPLEX~\cite{chung2024accurate} (three-resolution fusion) and M2OST~\cite{wang2025m2ost} (many-to-one Transformer);
    \item Generative modeling: STFlow~\cite{huang2025scalable} (flow matching).
\end{itemize}
To reduce differences attributable to image encoders, we replace the encoders of all compatible baselines with frozen UNI~\cite{chen2024towards}. ST-Net retains its native CNN due to its tightly coupled architecture. To quantify the performance gain from the encoder alone, the UNI-based baseline (frozen UNI encoder + two-layer MLP) is included. The results for plug-in variants and \method ablations are reported separately in Section~\ref{sec:stflowms_results} and Section~\ref{sec:ablation}.

\begin{table*}[t]
\centering
\caption{%
    PCC-200 results on nine HEST-1k datasets.
    Mean $\pm$ std across folds. Best in \textbf{bold}, second-best \underline{underlined}.
    $\dag$: \method with GO hierarchy.
    UNI denotes a frozen UNI encoder followed by a two-layer MLP. OOM denotes an out-of-memory error.
}
\label{tab:main}
\setlength{\tabcolsep}{4pt}
\begin{tabular}{lccccccccc}
\toprule
\textbf{Dataset} & \textbf{ST-Net} & \textbf{UNI} & \textbf{HisToGene} & \textbf{BLEEP} & \textbf{EGN} & \textbf{TRIPLEX} & \textbf{M2OST} & \textbf{STFlow} & \textbf{\method}$^\dag$ \\
\midrule
CCRCC      & 0.311{\small±0.087} & 0.353{\small±0.077} & 0.337{\small±0.107} & 0.334{\small±0.070} & 0.365{\small±0.074} & 0.420{\small±0.081} & 0.409{\small±0.092} & \textbf{0.437}{\small±0.056} & \underline{0.430}{\small±0.068} \\
COAD       & 0.412{\small±0.047} & 0.440{\small±0.114} & 0.535{\small±0.073} & 0.527{\small±0.061} & 0.519{\small±0.084} & 0.528{\small±0.076} & 0.512{\small±0.054} & \underline{0.538}{\small±0.094} & \textbf{0.546}{\small±0.104} \\
HCC        & 0.164{\small±0.095} & 0.179{\small±0.004} & 0.188{\small±0.059} & 0.170{\small±0.013} & 0.242{\small±0.039} & 0.222{\small±0.002} & 0.295{\small±0.037} & \underline{0.317}{\small±0.066} & \textbf{0.338}{\small±0.084} \\
IDC        & 0.589{\small±0.161} & 0.649{\small±0.070} & OOM                & 0.643{\small±0.073} & 0.673{\small±0.048} & 0.672{\small±0.051} & 0.691{\small±0.098} & \underline{0.702}{\small±0.056} & \textbf{0.705}{\small±0.057} \\
LUNG       & 0.572{\small±0.003} & 0.531{\small±0.012} & 0.590{\small±0.024} & 0.581{\small±0.017} & 0.574{\small±0.021} & 0.606{\small±0.026} & \textbf{0.642}{\small±0.001} & 0.623{\small±0.025} & \underline{0.624}{\small±0.026} \\
PRAD       & 0.393{\small±0.010} & 0.352{\small±0.024} & 0.365{\small±0.020} & 0.349{\small±0.084} & 0.398{\small±0.021} & 0.441{\small±0.013} & 0.435{\small±0.028} & \underline{0.444}{\small±0.012} & \textbf{0.444}{\small±0.002} \\
READ       & 0.273{\small±0.056} & 0.298{\small±0.001} & 0.330{\small±0.006} & 0.307{\small±0.083} & 0.289{\small±0.085} & \underline{0.358}{\small±0.101} & 0.351{\small±0.083} & 0.348{\small±0.053} & \textbf{0.392}{\small±0.048} \\
SKCM       & 0.689{\small±0.121} & 0.652{\small±0.050} & 0.701{\small±0.011} & 0.673{\small±0.008} & 0.744{\small±0.000} & 0.773{\small±0.015} & \underline{0.785}{\small±0.035} & 0.780{\small±0.002} & \textbf{0.793}{\small±0.016} \\
KIDNEY     & 0.319{\small±0.076} & 0.348{\small±0.047} & 0.330{\small±0.032} & 0.331{\small±0.065} & 0.374{\small±0.058} & \underline{0.378}{\small±0.041} & 0.365{\small±0.029} & 0.334{\small±0.035} & \textbf{0.385}{\small±0.048} \\
\midrule
\textbf{Avg} & 0.414 & 0.422 & N/A & 0.435 & 0.464 & 0.489 & 0.498 & \underline{0.503} & \textbf{0.517} \\
\bottomrule
\end{tabular}
\end{table*}

\subsubsection{Metrics.}
Let $\{\text{PCC}_g\}_{g=1}^G$ denote the per-gene Pearson correlation coefficients sorted in descending order. The evaluation metric PCC-$k$ is defined as $\text{PCC-}k = \frac{1}{k} \sum_{g=1}^{k} \text{PCC}_g$.
Results are reported for $k \in \{10, 50, 200\}$ in a log1p-transformed space. All values take the format \emph{mean $\pm$ standard deviation} across cross-validation folds.

\subsubsection{Implementation Details.}
The decoder of \method employs an embedding dimension of $d{=}512$, 4 attention heads, and $L{=}2$ shared Transformer layers. Both the dropout and drop-path rates are configured to 0.1. The Spatial Transformer encodes the neighborhood context utilizing the $n{=}8$ nearest spatial neighbors across $L_s{=}2$ layers. The training process utilizes the AdamW optimizer (with a learning rate of $10^{-4}$ and a weight decay of $10^{-4}$) coupled with a CosineAnnealingLR scheduler and an early stopping patience of 10 epochs. All experiments are implemented using PyTorch and run on a single NVIDIA RTX 5090 (32\,GB) GPU.

\subsection{Main Results (RQ1)}

Table~\ref{tab:main} presents the PCC-200 results across all nine datasets, and the additional PCC-10 and PCC-50 results are provided in Appendix~A. The proposed \method achieves a mean PCC-200 of 0.517, surpassing both the generative STFlow approach (0.503 on average) and all discriminative baselines. Despite not incorporating a generative process, \method outperforms STFlow on eight of nine datasets. The most substantial gains are observed on KIDNEY ($+0.051$), READ ($+0.044$), HCC ($+0.021$), and SKCM ($+0.013$), demonstrating that GO-based biological regularization provides robust structural guidance. STFlow retains an advantage only on CCRCC.

Among discriminative methods, \method surpasses M2OST and TRIPLEX, both multi-magnification baselines ($0.498$ and $0.489$ on average), by $+0.019$ and $+0.028$, respectively, and EGN, the strongest single-magnification regression baseline ($0.464$ on average), by $+0.053$. This gap is pronounced on HCC ($+0.116$ over TRIPLEX and $+0.174$ over ST-Net), confirming that GO-structured decoding delivers a substantial advantage on challenging, low-signal datasets. The UNI-based baseline achieves a modest average improvement over ST-Net (0.422 vs. 0.414), indicating the benefit of the frozen foundation encoder. The full \method framework improves over the UNI-based MLP baseline by $+0.095$, demonstrating the benefit of the proposed architecture beyond the frozen image encoder alone. HisToGene is excluded from the overall average comparison because it encountered an out-of-memory error on the TENX99 slide of the IDC dataset, whose spot count exceeds 20{,}000, causing slide-level self-attention to exceed the memory capacity of a 32\,GB GPU.

\begin{table}[t]
\small
\centering
\caption{%
    Plug-in compatibility results measured by PCC-200. The better result within each Base/+MS pair is shown in bold. 
}
\label{tab:plugin}
\setlength{\tabcolsep}{1.5pt}
\begin{tabular}{lccccccccc}
\toprule
& \multicolumn{3}{c}{\textbf{STFlow}} & \multicolumn{3}{c}{\textbf{EGN}} & \multicolumn{3}{c}{\textbf{ST-Net}} \\
\cmidrule(lr){2-4} \cmidrule(lr){5-7} \cmidrule(lr){8-10}
\textbf{Dataset} & Base & +MS & $\Delta$ & Base & +MS & $\Delta$ & Base & +MS & $\Delta$ \\
\midrule
CCRCC  & 0.437 & \textbf{0.451} & $+0.014$ & 0.365 & \textbf{0.371} & $+0.006$ & 0.311 & \textbf{0.319} & $+0.008$ \\
COAD   & 0.538 & \textbf{0.553} & $+0.015$ & 0.519 & \textbf{0.544} & $+0.025$ & 0.412 & \textbf{0.462} & $+0.050$ \\
HCC    & 0.317 & \textbf{0.337} & $+0.020$ & 0.242 & \textbf{0.272} & $+0.030$ & 0.164 & \textbf{0.190} & $+0.026$ \\
IDC    & 0.702 & \textbf{0.709} & $+0.007$ & 0.673 & \textbf{0.692} & $+0.019$ & 0.589 & \textbf{0.610} & $+0.021$ \\
LUNG   & \textbf{0.623} & 0.619 & $-0.004$ & 0.574 & \textbf{0.615} & $+0.041$ & 0.572 & \textbf{0.604} & $+0.032$ \\
PRAD   & 0.444 & \textbf{0.460} & $+0.016$ & \textbf{0.398} & 0.363 & $-0.035$ & 0.393 & \textbf{0.399} & $+0.006$ \\
READ   & 0.348 & \textbf{0.395} & $+0.047$ & 0.289 & \textbf{0.323} & $+0.034$ & \textbf{0.273} & 0.271 & $-0.002$ \\
SKCM   & 0.780 & \textbf{0.784} & $+0.004$ & 0.744 & \textbf{0.752} & $+0.008$ & 0.689 & \textbf{0.703} & $+0.014$ \\
KIDNEY & 0.334 & \textbf{0.386} & $+0.052$ & 0.374 & \textbf{0.400} & $+0.026$ & 0.319 & \textbf{0.326} & $+0.007$ \\
\midrule
\textbf{Avg} & 0.503 & \textbf{0.522} & $+0.019$ & 0.464 & \textbf{0.481} & $+0.017$ & 0.414 & \textbf{0.432} & $+0.018$ \\
\bottomrule
\end{tabular}
\end{table}

\begin{figure*}[htp!]
    \centering
    \includegraphics[width=0.99\linewidth]{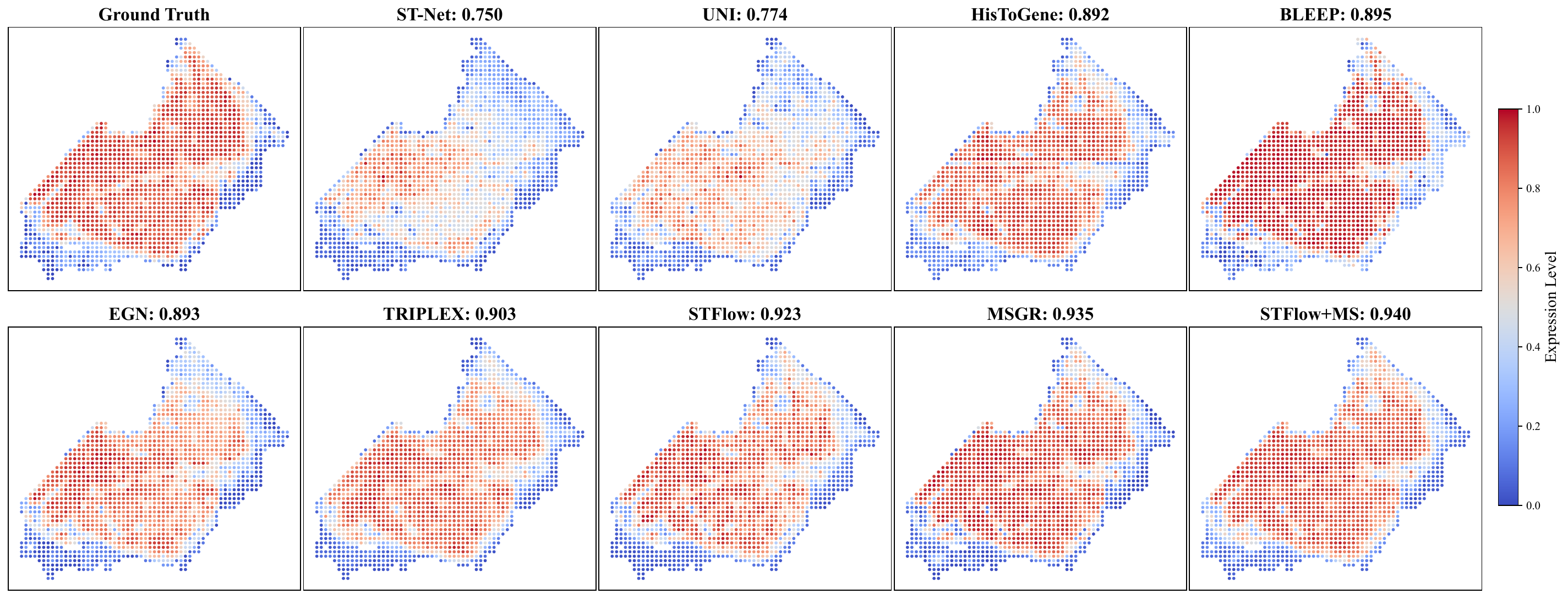}
    \caption{
        Predicted spatial expression of \textit{MLANA} on slide TENX115 from the SKCM dataset, compared against the ground truth.
        Each panel title reports the per-slide PCC.
        The color scale encodes the normalized expression level (blue: low, red: high).
    }
    \Description{Spatial heatmaps comparing ground-truth and predicted MLANA expression on slide TENX115 from the SKCM dataset. Panel titles report per-slide Pearson correlation coefficients, and color ranges from low blue to high red expression.}
    \label{fig:marker}
\end{figure*}

\subsection{Compatibility with Existing Methods (RQ2)}
\label{sec:stflowms_results}

To validate plug-in compatibility, the linear output heads of STFlow, EGN, and ST-Net are replaced with the proposed GO-guided decoder, yielding \stflowms, \egnms, and ST-Net+MS, respectively. The GO hierarchy is constructed offline once per dataset and shared across all cross-validation folds. For all three base models, modifications are strictly limited to: (i) replacing the linear output head with the GO-guided multi-scale decoder, and (ii) substituting the standard MSE loss with the multi-scale GO-supervised loss (Equation~\eqref{eq:loss}). Results are reported in Table~\ref{tab:plugin}.

The three variants improve average PCC-200 by $+0.019$, $+0.017$, and $+0.018$ for STFlow, EGN, and ST-Net, respectively, and each improves on eight of nine datasets. In particular, ST-Net+MS improves from $0.414$ to $0.432$ on average while retaining ST-Net's native DenseNet-121 encoder. Together, these results show that the GO-guided decoder is compatible with generative, exemplar-guided, and CNN-based architectures. Integration details and the computational overhead of the lightweight plug-in decoder are reported in Appendix~E.

\subsection{Visualization of Cancer Marker Genes (RQ3)}
\label{sec:marker_visualization}

The \textit{MLANA} gene encodes Melan-A (also known as MART-1), a melanocytic differentiation antigen commonly used as part of an immunohistochemical panel to support melanocytic lineage assignment in surgical pathology~\cite{ohsie2008immunohistochemical}. Spatial \textit{MLANA} expression therefore provides a biologically meaningful lineage-associated signal in the SKCM dataset. Accordingly, we use its spatial expression to evaluate the recovery of a localized molecular pattern.

As shown in Figure~\ref{fig:marker}, most methods recover the broad spatial expression pattern but differ in signal dispersion, local contrast, and boundary fidelity. ST-Net (PCC $=0.750$) and UNI ($0.774$) produce diffuse maps, whereas HisToGene ($0.892$) and EGN ($0.893$) recover the dominant high-expression region but underrepresent some local intensity variations. BLEEP ($0.895$) produces overly widespread high-expression signals. TRIPLEX ($0.910$) and STFlow ($0.923$) better capture the global pattern but still smooth out local contrast. In comparison, \method more closely reproduces both the global distribution and local variation, achieving a per-slide PCC of $0.935$. \stflowms achieves the highest PCC of $0.940$ ($+0.017$ over STFlow), with visually sharper expression transitions, suggesting that the GO-guided decoder improves local spatial fidelity in this example. While this example alone does not establish GO-guided coarse-to-fine refinement, Appendix~F.1 provides complementary evidence for the biological relevance of intermediate predictions by evaluating their alignment with Hallmark pathway activities.

\begin{table}[htp!]
\centering
\caption{%
    \textbf{Ablation study} (Mean PCC-200 across nine datasets).
    $\Delta$ denotes the performance difference relative to \method.
}
\label{tab:ablation}
\setlength{\tabcolsep}{4pt}
\begin{tabular}{lcc}
\toprule
\textbf{Model} & \textbf{Avg} & \textbf{$\Delta$} \\
\midrule
\textbf{\method} (GO hierarchy) & \textbf{0.517} & — \\
w/o latent highway & 0.502 & $-0.015$ \\
w/o residual inheritance & 0.504 & $-0.013$ \\
SSGT (single-scale, no hierarchy) & 0.497 & $-0.020$ \\
MSGR-Random (random hierarchy)    & 0.490 & $-0.027$ \\
\bottomrule
\end{tabular}
\end{table}

\subsection{Ablation Study (RQ4, RQ5)}
\label{sec:ablation}

To validate the key design choices of \method, five model configurations are compared across all nine datasets (Table~\ref{tab:ablation}): the full \method model, two variants that each remove one architectural component, the single-scale SSGT variant, and MSGR-Random.

\paragraph{Latent highway.}
The latent highway propagates latent context from the parent scale to each refinement step via cross-attention, enabling finer scales to directly access the biological context established at coarser scales. The removal of this component decreases the average PCC-200 from $0.517$ to $0.502$ ($-0.015$), confirming that cross-scale latent interaction is essential for coherent coarse-to-fine refinement. Without the latent highway, finer scales no longer receive parent-scale latent context through cross-attention, although they still inherit parent predictions through the residual pathway.

\paragraph{Residual inheritance.}
By default, \method predicts residual corrections relative to inherited parent-scale estimates. Replacing this mechanism with direct prediction at each scale (i.e., predicting $\hat{\mathbf{y}}^{(k)}$ directly without inheritance) reduces performance to $0.504$ ($-0.013$). By decomposing the learning problem into smaller, GO-aligned correction tasks, the residual formulation is easier to optimize than direct prediction of the full target values at every scale.

\paragraph{Multi-scale GO hierarchy vs.\ single-scale.}
The SSGT variant uses the same Transformer backbone as \method but predicts all 200 genes in a single step, without the GO hierarchy. The average PCC-200 of $0.497$ yields a difference of $-0.020$ relative to \method, confirming that multi-scale hierarchical refinement contributes meaningfully beyond what the backbone architecture alone provides. 

\paragraph{GO structure vs.\ random hierarchy.}
MSGR-Random replaces the GO tree with a randomly constructed hierarchy of identical shape while holding all other components constant. This variant achieves an average of $0.490$, trailing \method by $0.027$. It directly quantifies the contribution of the biologically grounded GO structure relative to a structurally equivalent but semantically arbitrary hierarchy. Appendix~C.2 further compares GO against pathway, PPI, coexpression, and random hierarchy priors under the same backbone.

\begin{figure}[t!]
\centering
\includegraphics[width=0.98\linewidth]{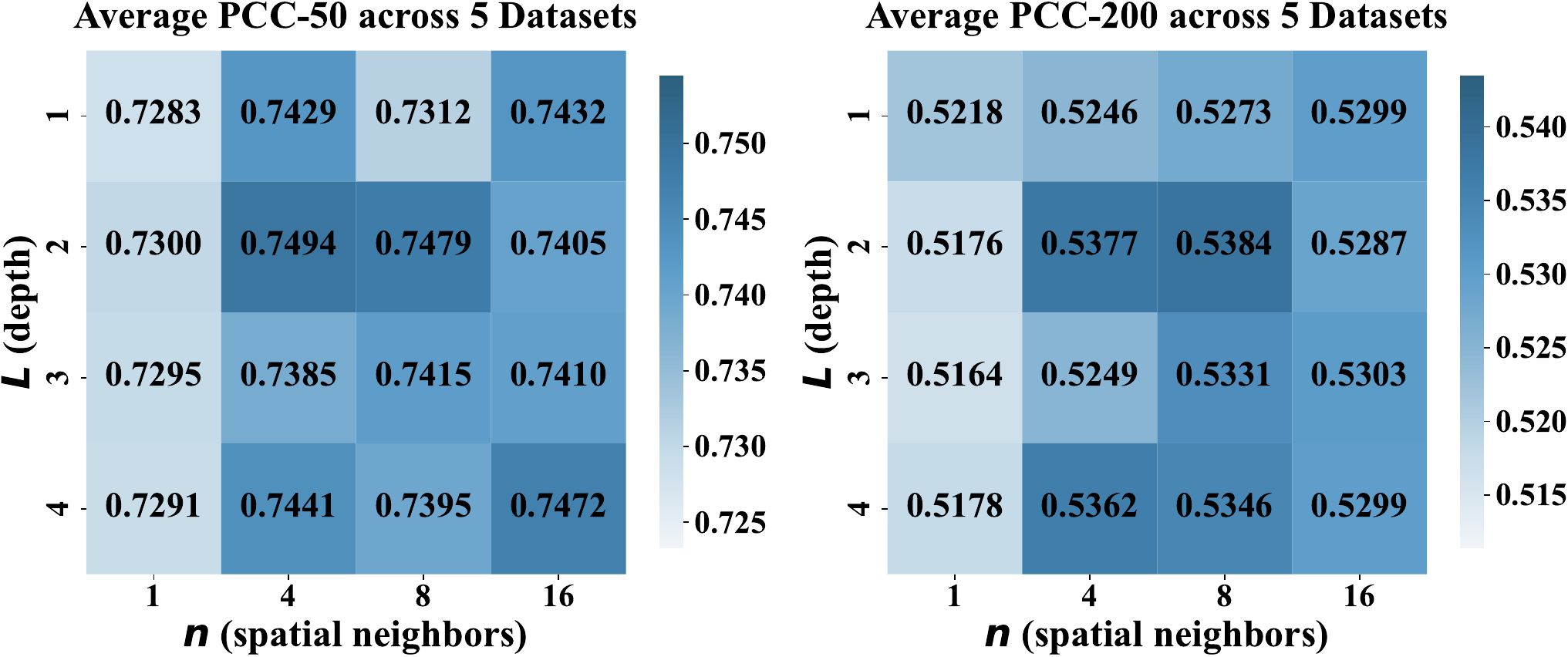}
\caption{%
    Hyperparameter sensitivity analysis of the proposed MSGR framework: average PCC-50 (left) and PCC-200 (right) across five datasets for all 16 $(n, L)$ configurations.
    The best-performing region for both metrics is $L{=}2$ with $n \in \{4, 8\}$.
}
\Description{Two heatmaps show average PCC-50 and PCC-200 across 16 combinations of spatial neighborhood size and shared Transformer depth. Both metrics are strongest at Transformer depth two with neighborhood size four or eight.}
\label{fig:hps_heatmap}
\end{figure}

\subsection{Hyperparameter Sensitivity (RQ6)}
\label{sec:hpsens}
Figure~\ref{fig:hps_heatmap} reports the average PCC-50 and PCC-200 scores across five datasets for all 16 combinations of spatial neighborhood size $n \in \{1,4,8,16\}$ and shared Transformer depth $L \in \{1,2,3,4\}$. Both metrics exhibit consistent sensitivity patterns: $n{=}1$ yields consistently suboptimal performance, whereas the best-performing region occurs at $L{=}2$ with $n \in \{4,8\}$. Among the nine configurations with at least moderate spatial aggregation and Transformer depth ($n \in \{4,8,16\}$ and $L \in \{2,3,4\}$), the ranges of PCC-50 and PCC-200 are only $0.011$ and $0.014$, respectively. This limited variation indicates that MSGR remains relatively robust within this practically relevant configuration range. Increasing $L$ from 1 to 2 improves both metrics for $n \in \{4,8\}$, whereas greater depths provide no consistent additional benefit. Moreover, performance at $n{=}16$ remains closer to the peak values than at $n{=}1$, suggesting greater tolerance to overly large neighborhoods than to overly small ones. Per-dataset results are provided in Appendix~D.

\section{Discussion}
\label{sec:discussion}

\paragraph{Why does GO-guided decoding work?}
The effectiveness of \method stems from two complementary mechanisms. First, hierarchical supervision provides a coarse-to-fine decomposition: hierarchy-level targets anchor each refinement step to biologically meaningful functional units, constraining the prediction space from coarse functional targets to gene-level outputs. Second, residual refinement decomposes the learning problem into increasingly fine-grained subtasks---beginning with a root-level estimate, followed by GO domain-level and GO term-level refinement, and ending with gene-level corrections. This progression follows the coarse-to-fine organization of biological concepts encoded by the GO hierarchy.

\paragraph{Complementarity to existing paradigms.}
The ablations reveal that removing either the latent highway or residual inheritance incurs comparable degradation ($-0.015$ and $-0.013$, respectively), indicating that cross-scale communication and residual correction serve complementary roles. Because the GO-guided decoder operates on the gene side independently of the image encoder, it can be integrated into diverse base models. Results from all three plug-in variants---\stflowms, \egnms, and ST-Net+MS---validate this compatibility: each improves performance on eight of nine datasets, with average PCC-200 gains of $+0.019$, $+0.017$, and $+0.018$, respectively. These results demonstrate that gene-side structural guidance complements diverse image-side modeling strategies.

\paragraph{Biological interpretability.}
Beyond predictive accuracy, the GO hierarchy enhances interpretability: predictions at intermediate scales correspond to biologically meaningful functional modules. Because clinically relevant markers frequently belong to specific pathways, the hierarchy enables the model to leverage pathway-level co-regulation, which conventional one-step decoding methods must infer implicitly from data. Quantitative alignment between intermediate GO-term predictions and pathway activities on SKCM is reported in Appendix~F.1. The limitation of mean aggregation for sparse pathway signals is discussed in Appendix~G.2.

\paragraph{Limitations and future work.}
\begin{enumerate}[leftmargin=*,noitemsep,topsep=2pt]
    \item GO annotations may be incomplete for less-studied genes. Consequently, fallback strategies such as clustering-based hierarchy construction could be employed to mitigate this limitation.
    \item The framework operates at the spot level, whereas modeling cell-type heterogeneity at sub-spot resolution remains an open direction for future research.
    \item The current design simplifies the GO DAG into a tree. However, preserving richer topological information, for example through ensemble decoding over multiple tree projections, may yield further improvements.
\end{enumerate}

\section{Conclusion}
\label{sec:conclusion}

We present \method, a framework that leverages the Gene Ontology hierarchy as a biological prior for coarse-to-fine spatial gene expression prediction from histopathology images. By organizing prediction across four GO-aligned scales with residual corrections and a shared Transformer backbone, \method consistently outperforms existing methods. Furthermore, the GO-guided decoder exhibits plug-in compatibility: integrating it into STFlow, EGN, and ST-Net yields average performance improvements across all three architectures. These results demonstrate the complementarity of GO-guided gene decoding and diverse image-side modeling strategies. Ablation studies confirm that the GO-derived hierarchy provides essential biological structure while cross-scale communication enables coherent coarse-to-fine refinement. Beyond accuracy, the hierarchical structure enhances interpretability by aligning intermediate predictions with biologically meaningful functional modules. More broadly, our findings highlight the potential of biological ontologies as structural priors for knowledge-guided biomedical prediction.

\begin{acks}
This work was supported by the Hunan Provincial Natural Science Foundation of China (No.~2026JJ60213).
\end{acks}

\bibliographystyle{ACM-Reference-Format}
\balance
\bibliography{msgr_refs}

\clearpage
\appendix

\setcounter{dbltopnumber}{4}    
\setcounter{topnumber}{4}        
\setcounter{totalnumber}{6}      
\renewcommand{\dbltopfraction}{0.95}
\renewcommand{\topfraction}{0.95}
\renewcommand{\textfraction}{0.05}
\renewcommand{\floatpagefraction}{0.85}


\section{Full Main Results}
\label{app:full_results}

Table~\ref{tab:datasets} summarizes the statistics of the nine spatial transcriptomics benchmarks utilized in this study.
Table~\ref{tab:pcc10} and Table~\ref{tab:pcc50} report the PCC-10 and PCC-50 metrics across all methods and datasets.
These metrics evaluate the performance on the top-10 and top-50 genes,
complementing the primary PCC-200 metric reported in the main paper.
Table~\ref{tab:full_metrics} provides a comprehensive comparison of all methods across five evaluation metrics. HisToGene is excluded from Table~\ref{tab:full_metrics} because it encountered out-of-memory errors on IDC (slide TENX99 exceeds 20,000 spots, which overflowed the slide-level self-attention on a 32\,GB GPU), thereby preventing the computation of the average across datasets.

\emph{Evaluation metrics.}
We adopt five metrics to comprehensively evaluate the quality of predictions, where $N$ denotes the number of spots, $G$ indicates the number of target genes, and $k$ represents the size of the gene subset. All metrics are computed between the predicted and ground-truth expression vectors for each test slide and are subsequently averaged across the cross-validation folds.

(i)~\textbf{PCC-}$k$. Let $\{\text{PCC}_g\}_{g=1}^G$ denote the per-gene Pearson correlation coefficients sorted in descending order. PCC-$k$ is defined as:
\begin{equation}
    \text{PCC-}k = \frac{1}{k} \sum_{g=1}^{k} \text{PCC}_g,
\end{equation}
where the per-gene Pearson correlation is:
\begin{equation}
    \text{PCC}_g = \frac{\textstyle\sum_{s}(y_{s,g} - \bar{y}_g)(\hat{y}_{s,g} - \bar{\hat{y}}_g)}{\sqrt{\textstyle\sum_{s}(y_{s,g} - \bar{y}_g)^2 \;\textstyle\sum_{s}(\hat{y}_{s,g} - \bar{\hat{y}}_g)^2}},
\end{equation}
and $y_{s,g}$, $\hat{y}_{s,g}$ represent the ground-truth and predicted expression levels of gene $g$ at spot $s$. We report PCC-10, PCC-50, and PCC-200.

(ii)~\textbf{MSE} (Mean Squared Error) measures the average squared deviation between the predicted and ground-truth expressions across all genes and spots within the log1p space:
\begin{equation}
    \text{MSE} = \frac{1}{NG} \sum_{s=1}^{N} \sum_{g=1}^{G} (y_{s,g} - \hat{y}_{s,g})^2.
\end{equation}

(iii)~\textbf{MAE} (Mean Absolute Error) provides an interpretable measure of error that is less sensitive to outliers:
\begin{equation}
    \text{MAE} = \frac{1}{NG} \sum_{s=1}^{N} \sum_{g=1}^{G} |y_{s,g} - \hat{y}_{s,g}|.
\end{equation}
For both MSE and MAE, lower values indicate superior prediction accuracy.

\emph{PCC-10 and PCC-50 analysis.}
Tables~\ref{tab:pcc10} and~\ref{tab:pcc50} reveal a consistent hierarchy of performance across gene subsets. Regarding PCC-10, \method achieves the highest average ($0.776$), surpassing both TRIPLEX ($0.763$) and STFlow ($0.763$) by $+0.013$. Regarding PCC-50, \method ($0.710$) extends the margin over STFlow ($0.694$) to $+0.016$. The margin remains substantial at PCC-200 ($+0.014$ over STFlow), confirming that decoding structured by GO provides consistent benefits across gene subsets of varying size, particularly on challenging datasets with low signals (HCC: $+0.036$ over STFlow on PCC-50, READ: $+0.029$). Notably, \method achieves the highest rank on five of the nine datasets for both PCC-10 and PCC-50 (excluding HisToGene, which encounters out-of-memory errors on IDC).

\emph{Comprehensive metric comparison.}
Table~\ref{tab:full_metrics} provides a holistic view across all five metrics. \method achieves the best PCC-10, PCC-50, PCC-200, and MAE ($0.865$), while STFlow attains the lowest MSE ($1.309$). The MSE of \method ($1.348$) is marginally higher than that of STFlow ($1.309$), which we attribute to the multi-scale supervision that regularizes intermediate predictions rather than optimizing solely for the MSE at the leaf level. This trade-off is favorable: the correlation gains ($+0.014$ in PCC-200) substantially outweigh the marginal increase in MSE ($+0.039$), as downstream biological analyses primarily rely on relative expression patterns rather than absolute magnitudes.


\begin{figure}[ht!]
\setlength{\abovecaptionskip}{0.2cm}
\centering
\includegraphics[width=0.95\linewidth]{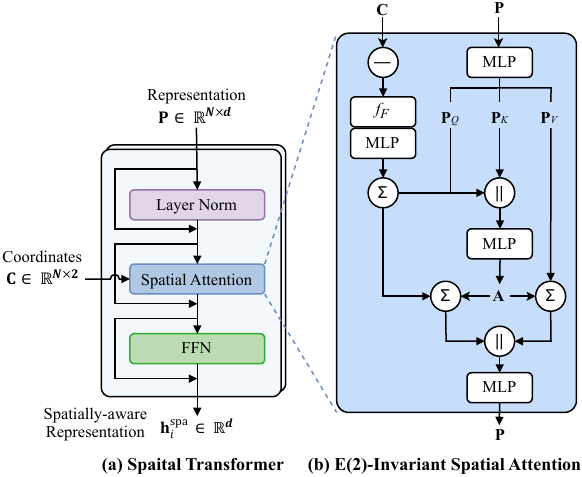}
\caption{
    \textbf{Spatial Transformer architecture.}
    (a)~Overall architecture.
    (b)~The E(2)-invariant spatial attention.
}
\Description{Spatial Transformer architecture diagram. Per-spot UNI features and neighboring spots are projected into tokens, processed by Transformer layers with E(2)-invariant frame averaging, and combined through spatial attention.}
\label{fig:spatial_transformer}
\end{figure}

\section{Spatial Transformer Architecture}
\label{app:spatial_transformer}

Figure~\ref{fig:spatial_transformer} illustrates the Spatial Transformer utilized within \method to encode context from the neighborhood. This module builds upon the transformer backbone of STFlow~\cite{huang2025scalable}. Given the UNI embeddings of a center spot and its $n$ nearest spatial neighbors, the module applies $L_s$ Transformer layers to aggregate information from the neighborhood while preserving spatial equivariance.

\emph{E(2)-equivariant frame averaging.}
To ensure that the spatial encoding is invariant to rotations and reflections of the tissue structure, we employ E(2)-equivariant frame averaging~\cite{puny2021frame}. For each center spot $i$ and its neighbor $j$, we compute the relative coordinate $\mathbf{r}_{ij} = \mathbf{c}_j - \mathbf{c}_i$ and construct a local covariance matrix:
\begin{equation}
    \Sigma_i = \sum_{j \in \mathcal{N}(i)} \mathbf{r}_{ij}\mathbf{r}_{ij}^\top.
\end{equation}
Eigendecomposing $\Sigma_i$ yields an orthonormal basis:
\begin{equation}
    \Sigma_i = U \Lambda U^\top,
\end{equation}
where $U \in \mathbb{R}^{d \times d}$ is the matrix of eigenvectors. By further enumerating sign flips across all $2^d = 4$ axis combinations (for $d{=}2$ spatial dimensions), we obtain a complete set of frames:
\begin{equation}
    R_g = U \cdot \mathrm{diag}(s_g), \quad s_g \in \{-1, +1\}^d, \quad g = 1, \dots, 4,
\end{equation}
that cover all rotations and reflections. For each frame $R_g$, the edge embedding is computed as:
\begin{equation}
    \mathbf{e}_{ij}^{(g)} = \phi \bigl(R_g \mathbf{r}_{ij},\; \|\mathbf{r}_{ij}\|\bigr),
\end{equation}
where $\phi$ is an MLP encoding the rotated relative coordinates and the inter-spot distance. The final edge embedding is obtained by averaging across all frames:
\begin{equation}
    \mathbf{e}_{ij} = \frac{1}{4} \sum_{g=1}^{4} \mathbf{e}_{ij}^{(g)}.
\end{equation}
This frame averaging guarantees exact E(2)-equivariance of the spatial encoding.

\emph{MLP-based spatial attention.}
Unlike dot-product attention, our spatial attention employs a Multilayer Perceptron (MLP) to compute attention weights from concatenated query, key, and edge features. Specifically, for a center spot $i$ with neighbor $j$, the attention input comprises: (i)~the query embedding $\mathbf{q}_i$, (ii)~the key embedding $\mathbf{k}_j$, and (iii)~the edge embedding $\mathbf{e}_{ij}$ derived from the frame-averaged relative coordinates and the distance between spots. A shared MLP maps this concatenated feature to a scalar attention logit:
\begin{equation}
    \alpha_{ij} = \frac{\exp \bigl(f(\mathbf{q}_i \,\|\, \mathbf{k}_j \,\|\, \mathbf{e}_{ij})\bigr)}{\sum_{j' \in \mathcal{N}(i)} \exp \bigl(f(\mathbf{q}_i \,\|\, \mathbf{k}_{j'} \,\|\, \mathbf{e}_{ij'})\bigr)},
\end{equation}
where $f$ denotes the shared MLP, $\|$ indicates concatenation, and $\mathcal{N}(i)$ is the set of $n$ nearest neighbors of spot $i$. The aggregated representation is then computed by separately weighting the value and edge streams, concatenating, and passing through an output MLP:
\begin{equation}
    \mathbf{h}^{\text{spa}}_i = \mathrm{MLP}_o \biggl(\Bigl(\sum_{j \in \mathcal{N}(i)} \alpha_{ij} \mathbf{v}_j\Bigr) \,\Big\|\, \Bigl(\sum_{j \in \mathcal{N}(i)} \alpha_{ij} \mathbf{e}_{ij}\Bigr)\biggr),
\end{equation}
where $\mathbf{v}_j$ is the value projection of neighbor $j$ and $\mathrm{MLP}_o$ is the output projection. This design allows the attention mechanism to explicitly condition on geometric relationships while remaining efficient in terms of parameters.


\section{Extended Ablation Studies}
\label{app:ablation}

\subsection{Multi-Scale Loss Weights Ablation}

To validate the design of multi-scale loss weights within the architecture of \method, we conducted targeted experiments on five core datasets (COAD, HCC, READ, LUNG, SKCM). Table~\ref{tab:extend_ablation} reports the results relative to the adopted configuration.

Uniform weighting across all scales (loss=[1,1,1,1]) degrades the performance by $-0.0144$, confirming that treating all levels of the hierarchy equally is suboptimal. Adding progressive weighting while retaining supervision at the root (loss=[1,1,2,4]) slightly narrows the performance gap to $-0.0137$ compared to uniform weighting, indicating that progressive emphasis on finer scales provides limited benefit when noise at the root level remains. Further amplifying the progressive weighting (loss=[1,2,4,8]) implicitly suppresses root supervision by assigning it the smallest weight relative to finer scales, substantially reducing the gap ($-0.0057$). This demonstrates that the virtual root node aggregates all $G$ genes into a single noisy target of global mean, and its gradient signal conflicts with objectives at finer scales. The optimal configuration loss=[0,1,2,4] eliminates this noisy supervision at the root while preserving the progressive emphasis toward predictions at the gene level, thereby achieving the optimal balance between multi-scale regularization and the signal-to-noise ratio. Notably, the entire performance variation across all tested configurations is only $0.0144$ in PCC-200, demonstrating that \method is robust to the specific choice of loss weights while still benefiting from removing supervision at the virtual root and progressively emphasizing finer scales.

\subsection{Biological-Prior Ablation}

To isolate the contribution of different biological priors within the same backbone of MSGR, we compare five strategies for hierarchy construction: \textbf{GO} (default), \textbf{Random}, \textbf{Pathway}, \textbf{PPI}, and \textbf{Coexp}.

\emph{Construction logic of the four alternative priors.}
All four alternatives are built from the same 200 target genes per dataset and are subsequently converted into hierarchical supervision targets:
(i)~\textbf{Random}: replaces the GO-derived tree with a randomly constructed hierarchy of identical shape while maintaining all other components as constant, thereby removing biological semantics while preserving the multi-scale decoding interface.
(ii)~\textbf{Pathway}: gene sets are sourced from the Reactome pathway database~\cite{gillespie2022reactome} via Enrichr~\cite{kuleshov2016enrichr}, filtered and selected by coverage; each gene is assigned to one representative set; unmatched genes are placed in an isolated set, yielding a compact three-level structure (root$\rightarrow$set$\rightarrow$gene).
(iii)~\textbf{PPI}: we construct a weighted gene graph from STRING~\cite{szklarczyk2025string} interactions, split it by connected components, and then partition each component into modules via community detection (with the merging of small modules).
(iv)~\textbf{Coexp}: correlations between genes are computed from ST expression profiles in the training set, edges are sparsified by per-gene top-$k$ neighbor selection combined with a correlation threshold, and then the same component and module partitioning is applied with a fallback for isolated genes.

\emph{Quantitative comparison.}
Table~\ref{tab:bio_prior_ablation} reports the averages across nine datasets. GO achieves the best results across all five metrics, confirming its effectiveness as a biological prior. Among the three biological alternatives, Pathway achieves the highest rank on PCC-200 ($0.506$) while Coexp attains the lowest MSE ($1.394$); all three are closely matched on PCC-10 ($0.762$--$0.768$) and PCC-50 ($0.695$--$0.699$).

At the per-dataset level (PCC-200), GO ranks first on six of the nine datasets, Pathway on two of the nine, and PPI on one of the nine (Random and Coexp: 0/9). Importantly, MSGR-RANDOM performs worse than MSGR-GO by $0.027$ on average for PCC-200, supporting the claim that improvements derive from the \emph{biological ontology structure} rather than the hierarchy decomposition alone.


\section{Per-Dataset Hyperparameter Sensitivity}
\label{app:hps_datasets}

\begin{figure*}[t!]
\centering
\includegraphics[width=0.95\linewidth]{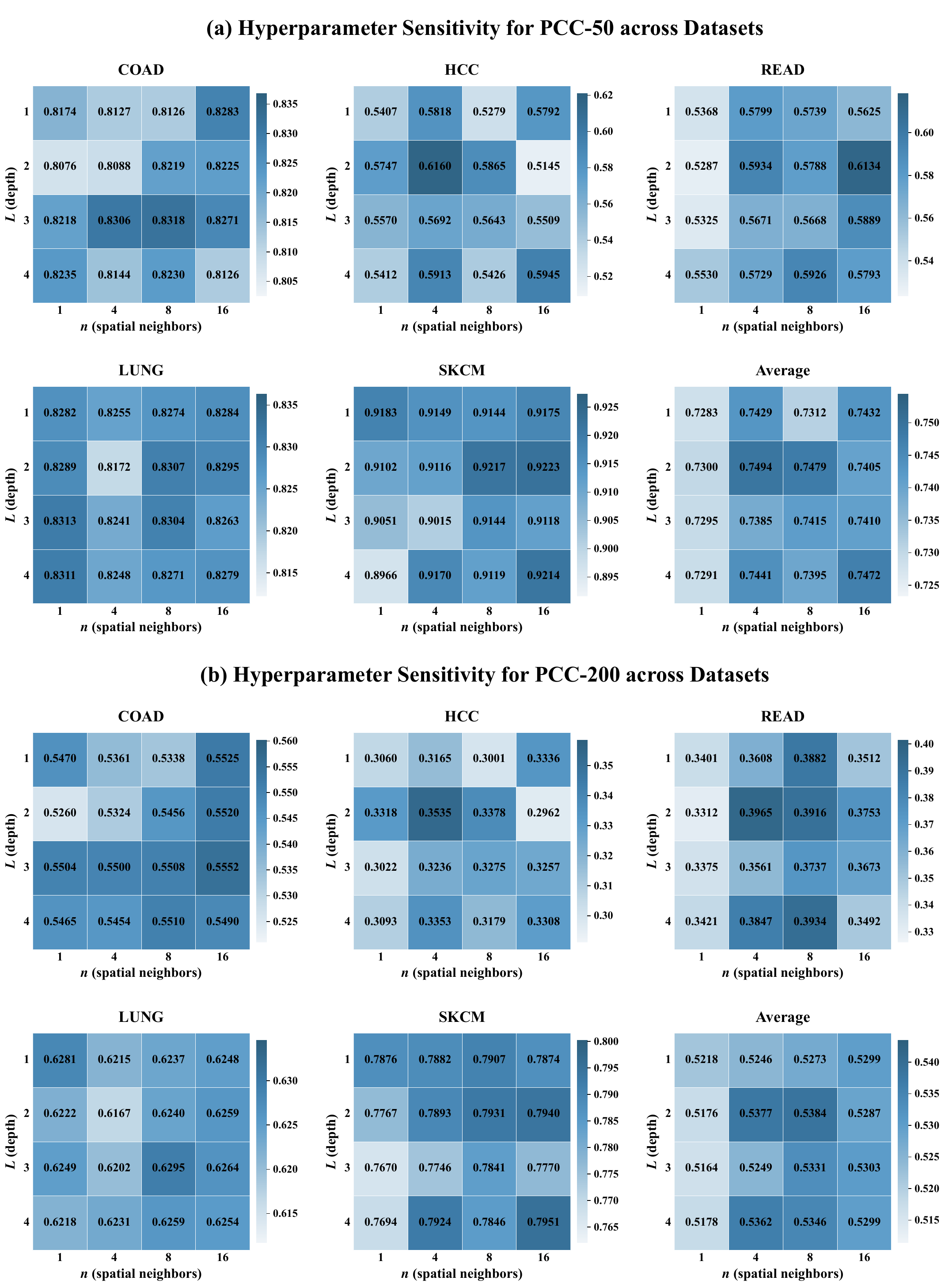}
\caption{%
    \textbf{Per-dataset hyperparameter sensitivity across five evaluation datasets.}
    Full $4{\times}4$ grids over spatial neighborhood size $n$ and Transformer depth $L$, with the average shown in the final panel. (a)~PCC-50. (b)~PCC-200. Cell values report the corresponding score, and each dataset uses an independent color scale.
}
\Description{Two grids of heatmaps showing hyperparameter sensitivity for five datasets and their average. Each heatmap covers 16 combinations of spatial neighborhood size and Transformer depth. Panel (a) reports PCC-50 and panel (b) reports PCC-200; each dataset has its own color scale.}
\label{fig:hps_all}
\end{figure*}

Figure~\ref{fig:hps_all} shows the full $4{\times}4$ $(n, L)$ sensitivity grid for each of the five evaluation datasets alongside the average: panel~(a) reports PCC-50 and panel~(b) reports PCC-200.

\emph{PCC-50 (Figure~\ref{fig:hps_all}(a)).}
The PCC-50 grid reveals amplified dynamic ranges on the most challenging datasets. HCC exhibits the widest range ($0.101$), with the optimal configuration at $(n{=}4, L{=}2)$ extending the margin over the $n{=}1$ column by $+0.063$. READ similarly shows a wide range of $0.085$, with the optimal configuration at $(n{=}16, L{=}2)$. In contrast, LUNG remains highly stable (range $0.014$). The average PCC-50 peaks at $(n{=}4, L{=}2)$ with $0.749$.

\emph{PCC-200 (Figure~\ref{fig:hps_all}(b)).}
HCC and READ exhibit the widest dynamic ranges ($0.057$ and $0.065$, respectively): their optimal configurations reside in the plateau ($n \in \{4,8\}$, $L{=}2$) while the $n{=}1$ column drops sharply. LUNG is the most stable tissue, which is consistent with its spatially homogeneous expression patterns. Three of the five datasets (HCC, READ, SKCM) have their lowest column mean at $n{=}1$; COAD and LUNG show comparable values at $n{=}1$ and $n{=}4$ (within $0.004$), indicating that the benefit of aggregating spatial information is smaller for tissues with an already high baseline PCC or spatially uniform expression. The average PCC-200 peaks at $(n{=}8, L{=}2)$ with $0.538$. The distinct average optima align with the observation in the main text that predicting fewer genes benefits from a more restricted spatial focus while larger gene sets require a broader context.

\section{Details of Integrating the GO-Guided Decoder into Existing Methods}
\label{app:integration}

The GO-guided decoder is designed as a model-agnostic module that can be integrated into existing architectures with minimal modifications. For STFlow~\cite{huang2025scalable}, EGN~\cite{yang2023exemplar}, and ST-Net~\cite{he2020integrating}, the integration follows a consistent pattern: (i)~the original linear output head is replaced with the hierarchical decoder, and (ii)~the standard MSE loss is substituted with the multi-scale GO-supervised loss. The GO hierarchy is constructed offline once per dataset and is shared across all cross-validation folds, introducing no additional data preprocessing overhead during training.

\emph{Architecture-specific integration.}
While the pattern of integration is consistent, the specific architectural context differs across methods. For STFlow, the decoder replaces the final linear head of the flow-matching denoiser while preserving the E(2)-equivariant spatial encoder. For EGN, the decoder substitutes the output projection after the exemplar-guided Transformer, maintaining the cross-attention mechanism for interaction with reference samples. For ST-Net, the hierarchical decoder directly follows the DenseNet-121 backbone~\cite{huang2017densely}, replacing the original single-layer output head. This architectural diversity, spanning generative flow-matching, exemplar-guided regression, and CNN-based encoding, demonstrates that the GO-guided decoder operates orthogonally to the paradigm of image-side feature extraction. Notably, the plug-in gains are consistent across image encoders: under ST-Net's native DenseNet-121 encoder the GO-guided decoder improves PCC-200 by $+0.018$ on average (Table~\ref{tab:plugin}), matching the $+0.017$ and $+0.019$ gains obtained under the frozen UNI encoder for EGN and STFlow. This indicates that the benefit of the GO-guided decoder is largely encoder-agnostic.

\emph{Lightweight MLP-based decoder.}
To ensure efficiency in parameters when integrating with existing methods, we employ a lightweight variant of the GO-guided decoder based on MLP layers rather than the full Transformer backbone used in MSGR. Specifically, each scale uses a two-layer MLP with SiLU activation for residual prediction, and cross-scale context is propagated via simple concatenation rather than cross-attention. This design maintains a minimal parameter overhead while still utilizing the structure of the GO hierarchy. Despite its reduced capacity, the lightweight decoder consistently improves the performance across base architectures, demonstrating that the gain primarily stems from the biological prior rather than an increased capacity of the model.

\emph{Computational overhead.}
For the lightweight MLP-based decoder used in the plug-in variants, Table~\ref{tab:decoder_overhead} compares the GO-guided head with the original flat MLP head. Across EGN and ST-Net, the GO-guided decoder adds 0.25\%--0.30\% FLOPs, 5.7\%--6.0\% training time per step, and 1.5\%--6.2\% GPU memory. The GO hierarchy itself is constructed offline once per dataset and introduces no additional runtime preprocessing cost during training.


\section{Additional Experiments}
\label{app:additional_experiments}

\subsection{Quantitative Validation of Intermediate-Scale Interpretability}

MSGR exposes predictions for named GO terms at intermediate levels. To quantify pathway-level signals captured by these predictions, we evaluate their alignment with MSigDB Hallmark pathway activities on SKCM. A GO-term activity is defined as MSGR's prediction for the corresponding node, whereas a Hallmark pathway activity is the mean ground-truth expression of pathway genes overlapping the 200-gene panel. We match GO terms to Hallmark pathways by Jaccard similarity, retaining pairs with at least two overlapping genes and the top three pathways per GO term. Pearson correlation is then computed across all spots for each matched pair.

This procedure yields 20 matched pairs with a mean correlation of $r{=}0.896$. Table~\ref{tab:msigdb_interpretability} reports the top 10 pairs, together with their biological implications in SKCM. The gene sets used to form each pair differ substantially in size, with a median overlap ratio of 13.2\% across all 20 pairs. All top-matched pathways are cancer-related; for example, EMT reflects the phenotype switching that drives invasion and metastasis in SKCM~\cite{tang2020emt}, and KRAS Signaling Up reflects RAS/MAPK activation, a central oncogenic driver of melanoma~\cite{cancer2015genomic}. These results indicate that the intermediate GO-term predictions are aligned with biologically meaningful pathway-level variation beyond individual gene outputs.

\subsection{Single-Parent versus Multi-Parent GO Projection}

The default hierarchy assigns each gene to its most specific GO term, yielding a single-parent tree. We compare this design with a multi-parent averaging variant (MSGR-MP), in which each gene inherits the average prediction of all of its GO-annotated parent terms before its residual correction is added. Other components are held fixed. Table~\ref{tab:multiparent_projection} shows that the single-parent tree outperforms the multi-parent averaging variant on all nine datasets, with an average PCC-200 difference of $-0.017$ for MSGR-MP. This experiment restores only multi-parent gene-to-term annotations; it neither retains nor restores the additional term-to-term connectivity of the native GO DAG. Thus, it evaluates uniform averaging over alternative parent-term predictions rather than a full DAG-aware decoder. The lower performance is consistent with such averaging diluting the more specific functional signal selected by the default projection: in the native GO DAG, multi-parent genes are annotated to between 2 and 14 terms (7.3 on average), and uniform averaging over these alternative parent terms can attenuate the discriminative signal carried by the most specific term.

\subsection{Structure of the Constructed GO Hierarchy}
\label{app:go_hierarchy_structure}

\begin{figure*}[ht!]
\centering
\includegraphics[width=0.9\linewidth]{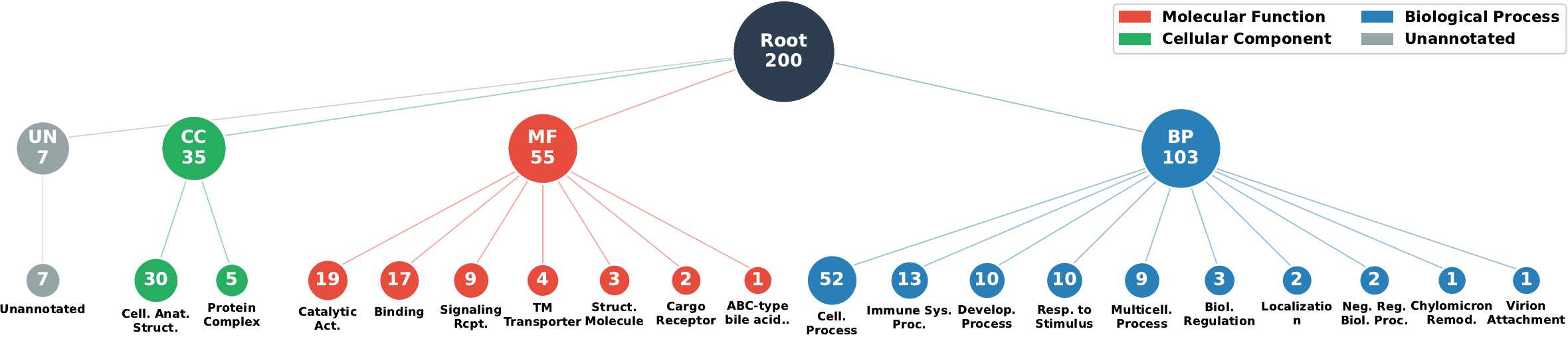}
\caption{GO hierarchy constructed for the IDC dataset. Node size is proportional to the number of annotated genes. Four Level-1 domains (MF, CC, BP, and an unannotated proxy UN) contain 20 Level-2 GO terms, with 200 leaf genes distributed according to annotation specificity.}
\Description{A tree diagram of the GO hierarchy constructed for the IDC dataset, showing a virtual root, four domain nodes (molecular function, cellular component, biological process, and an unannotated proxy), 20 GO-term nodes, and 200 gene leaves. Node size is proportional to the number of annotated genes.}
\label{fig:go_hierarchy}
\end{figure*}

To make the constructed hierarchy concrete, Figure~\ref{fig:go_hierarchy} visualizes the tree built for the IDC dataset. Consistent with the four-level design in Section~\ref{sec:go_hier}, the hierarchy contains one virtual root, four Level-1 domain nodes---molecular function (MF), cellular component (CC), biological process (BP), and an unannotated proxy (UN)---and 20 Level-2 GO terms, with the 200 target genes distributed across these terms according to GO annotation specificity.

A small number of Level-2 GO terms are singletons that annotate only one gene (3 of 20 in IDC; for instance, the term ``chylomicron remodeling'' is associated with a single gene). For such nodes the residual-correction mechanism has a naturally limited role, since the parent target and its single child carry nearly identical information. These singleton terms correspond to highly specific biological functions, and the path-compression strategy (Section~\ref{sec:go_hier}) already removes longer single-child chains, so they do not introduce additional depth into the hierarchy.

\section{Methodological Clarifications}
\label{app:methodological_clarifications}

\subsection{Residual Inheritance and the Zero-Sum Property}
\label{app:zero_sum}

Residual inheritance does not impose an explicit zero-sum constraint on the predicted corrections. Consider a parent node $p$ with children $c \in \mathcal{C}(p)$ and let $n_c = |\mathrm{desc}(c)|$ denote the number of descendant leaf genes. Because each parent target is the mean of its descendant leaf targets, the targets satisfy
\begin{equation}
    \sum_{c \in \mathcal{C}(p)} n_c\bigl(\mathbf{y}_c - \mathbf{y}_p\bigr) = \mathbf{0}.
    \label{eq:weighted_zero_sum}
\end{equation}
Consequently, if the parent and child targets are fitted exactly and the ReLU activation does not clip the output, the ideal child corrections obey a \emph{subtree-size-weighted} relation. This differs from the unweighted condition $\sum_c \boldsymbol{\Delta}_c = \mathbf{0}$ and is not enforced by either the architecture or the loss. During training, each scale minimizes its own MSE objective, while parent predictions and child corrections are jointly optimized; therefore, no zero-sum equality is imposed on the predicted corrections. Moreover, self-attention operates jointly over all nodes at a scale before the shared output head produces residuals, allowing sibling representations to coordinate. The main-paper residual-inheritance ablation decreases PCC-200 by $0.013$, empirically supporting this refinement mechanism.

\subsection{Mean Aggregation of Descendant Genes}
\label{app:mean_aggregation}

The coarse-scale target is the mean expression of a node's descendant genes. This aggregation can attenuate a pathway signal driven by only a small subset of its genes, and it is therefore not intended to preserve every sparse gene-specific effect at the intermediate scale. Instead, it supplies a biologically aligned auxiliary target, while the leaf level receives the largest loss weight ($w_3=4$) and remains individually supervised. The quantitative analysis in Section~F.1 provides empirical support that dominant pathway-level variation is retained: intermediate GO-term predictions align with independent MSigDB Hallmark activities on SKCM. More selective aggregation functions, such as variance-weighted or attention-based pooling, remain a direction for future work.

\begin{table*}[h!]
\centering
\caption{Computational overhead of the lightweight GO-guided decoder relative to a flat MLP head.}
\label{tab:decoder_overhead}
\setlength{\tabcolsep}{5pt}
\begin{tabular}{lcccccc}
\toprule
& \multicolumn{2}{c}{\textbf{FLOPs (M)}} & \multicolumn{2}{c}{\textbf{Training Time (ms/step)}} & \multicolumn{2}{c}{\textbf{GPU Memory (MB)}} \\
\cmidrule(lr){2-3}\cmidrule(lr){4-5}\cmidrule(lr){6-7}
\textbf{Model} & \textbf{Flat} & \textbf{GO-guided} & \textbf{Flat} & \textbf{GO-guided} & \textbf{Flat} & \textbf{GO-guided} \\
\midrule
EGN & 3500.3 & 3509.0 ($+0.25\%$) & 52.8 & 56.0 ($+6.0\%$) & 5145 & 5465 ($+6.2\%$) \\
ST-Net & 2864.7 & 2873.4 ($+0.30\%$) & 81.9 & 86.6 ($+5.7\%$) & 8870 & 9005 ($+1.5\%$) \\
\bottomrule
\end{tabular}
\end{table*}

\begin{table*}[h!]
\centering
\caption{%
    \textbf{Dataset statistics for the nine spatial transcriptomics benchmarks.}
    All datasets are sourced from HEST-1k~\cite{jaume2024hest}.
    \label{tab:datasets}
}
\setlength{\tabcolsep}{3pt}
\begin{tabular}{lccccccccc}
\toprule
 & \textbf{IDC} & \textbf{PRAD} & \textbf{SKCM} & \textbf{COAD} & \textbf{READ} & \textbf{CCRCC} & \textbf{HCC} & \textbf{LUNG} & \textbf{KIDNEY} \\
\midrule
Organ       & Breast   & Prostate & Skin    & Colon  & Rectum & KIDNEY  & Liver  & Lung   & KIDNEY \\
Technology  & Xenium   & Visium   & Xenium  & Xenium & Visium & Visium  & Visium & Xenium & Visium \\
\#Slides    & 4        & 23       & 2       & 4      & 4      & 24      & 2      & 2      & 23     \\
\#Patients  & 4        & 2        & 2       & 2      & 2      & 24      & 2      & 2      & 22     \\
\#Splits    & 4        & 2        & 2       & 2      & 2      & 6       & 2      & 2      & 4      \\
Avg.\ spots & 8884     & 2717     & 1517    & 3913   & 2102   & 3093    & 2098   & 2603   & 1122   \\
\bottomrule
\end{tabular}
\end{table*}

\begin{table*}[h!]
\centering
\caption{%
    \textbf{PCC-10 on 9 HEST-1k datasets.}
    Mean across folds. Best in \textbf{bold}, second-best \underline{underlined}. OOM denotes an out-of-memory error. \dag: \method with GO hierarchy.
    \label{tab:pcc10}
}
\setlength{\tabcolsep}{4pt}
\begin{tabular}{lcccccccc}
\toprule
\textbf{Dataset} & ST-Net & UNI & HisToGene & BLEEP & EGN & TRIPLEX & STFlow & \textbf{\method}$^\dag$ \\
\midrule
CCRCC      & 0.621 & 0.637 & 0.626 & 0.626 & 0.671 & \textbf{0.746} & 0.716 & \underline{0.719} \\
COAD       & 0.729 & 0.792 & \textbf{0.897} & 0.879 & 0.888 & 0.865 & 0.880 & \underline{0.888} \\
HCC        & 0.465 & 0.474 & 0.450 & 0.442 & 0.591 & 0.523 & \underline{0.655} & \textbf{0.679} \\
IDC        & 0.815 & 0.857 & OOM   & 0.878 & 0.890 & 0.882 & \underline{0.904} & \textbf{0.909} \\
LUNG       & 0.869 & 0.804 & 0.857 & 0.860 & 0.882 & \textbf{0.901} & 0.887 & \underline{0.893} \\
PRAD       & 0.591 & 0.552 & 0.547 & 0.561 & 0.606 & \textbf{0.671} & \underline{0.651} & 0.640 \\
READ       & 0.529 & 0.592 & 0.574 & 0.595 & 0.620 & \underline{0.654} & 0.641 & \textbf{0.663} \\
SKCM       & 0.891 & 0.877 & 0.906 & 0.894 & 0.924 & 0.926 & \underline{0.938} & \textbf{0.945} \\
KIDNEY     & 0.591 & 0.604 & 0.524 & 0.580 & 0.624 & \textbf{0.695} & 0.591 & \underline{0.650} \\
\midrule
\textbf{Avg} & 0.678 & 0.688 & / & 0.702 & 0.744 & \underline{0.763} & \underline{0.763} & \textbf{0.776} \\
\bottomrule
\end{tabular}
\end{table*}

\begin{table*}[h!]
\centering
\caption{%
    \textbf{PCC-50 on 9 HEST-1k datasets.}
    Mean across folds. Best in \textbf{bold}, second-best \underline{underlined}. OOM denotes an out-of-memory error. \dag: \method with GO hierarchy.
    \label{tab:pcc50}
}
\setlength{\tabcolsep}{4pt}
\begin{tabular}{lcccccccc}
\toprule
\textbf{Dataset} & ST-Net & UNI & HisToGene & BLEEP & EGN & TRIPLEX & STFlow & \textbf{\method}$^\dag$ \\
\midrule
CCRCC      & 0.532 & 0.550 & 0.536 & 0.533 & 0.577 & \textbf{0.655} & 0.634 & \underline{0.636} \\
COAD       & 0.648 & 0.700 & \textbf{0.826} & 0.806 & 0.814 & 0.795 & 0.816 & \underline{0.822} \\
HCC        & 0.340 & 0.361 & 0.375 & 0.358 & 0.475 & 0.414 & \underline{0.551} & \textbf{0.587} \\
IDC        & 0.778 & 0.817 & OOM   & 0.843 & 0.858 & 0.854 & \underline{0.876} & \textbf{0.880} \\
LUNG       & 0.799 & 0.747 & 0.801 & 0.798 & 0.822 & \textbf{0.832} & 0.828 & \underline{0.831} \\
PRAD       & 0.537 & 0.496 & 0.498 & 0.494 & 0.543 & \textbf{0.607} & \underline{0.593} & 0.585 \\
READ       & 0.443 & 0.490 & 0.491 & 0.496 & 0.522 & \underline{0.563} & 0.550 & \textbf{0.579} \\
SKCM       & 0.863 & 0.831 & 0.873 & 0.855 & 0.897 & 0.899 & \underline{0.912} & \textbf{0.922} \\
KIDNEY     & 0.482 & 0.503 & 0.449 & 0.478 & 0.531 & \textbf{0.562} & 0.484 & \underline{0.547} \\
\midrule
\textbf{Avg} & 0.602 & 0.611 & N/A & 0.629 & 0.671 & 0.687 & \underline{0.694} & \textbf{0.710} \\
\bottomrule
\end{tabular}
\end{table*}

\begin{table*}[h!]
\centering
\caption{%
    \textbf{Comprehensive comparison of all methods on 9 HEST-1k datasets.}
    Mean PCC-10, PCC-50, PCC-200, MSE, and MAE across all 9 datasets.
    Best in \textbf{bold}, second-best \underline{underlined}. \dag: \method with GO hierarchy.
    \label{tab:full_metrics}
}
\setlength{\tabcolsep}{4pt}
\begin{tabular}{lccccc}
\toprule
\textbf{Method} & \textbf{PCC-10} & \textbf{PCC-50} & \textbf{PCC-200} & \textbf{MSE}$\downarrow$ & \textbf{MAE}$\downarrow$ \\
\midrule
ST-Net        & 0.678 & 0.602 & 0.414 & 1.583 & 0.946 \\
UNI           & 0.688 & 0.611 & 0.422 & 1.463 & 0.920 \\
BLEEP         & 0.702 & 0.629 & 0.435 & 1.466 & 0.910 \\
EGN           & 0.744 & 0.671 & 0.464 & 1.460 & 0.904 \\
TRIPLEX       & \underline{0.763} & 0.687 & 0.489 & \underline{1.338} & 0.876 \\
STFlow        & \underline{0.763} & \underline{0.694} & \underline{0.503} & \textbf{1.309} & \underline{0.867} \\
\textbf{\method}$^\dag$      & \textbf{0.776} & \textbf{0.710} & \textbf{0.517} & 1.348 & \textbf{0.865} \\
\bottomrule
\end{tabular}
\end{table*}

\begin{table*}[h!]
\centering
\caption{%
    \textbf{Ablation results on loss weight design} (5-dataset mean PCC-200).
    [0,1,2,4] (bold) is the adopted configuration.
    $\Delta$ = difference from the adopted configuration.
}
\label{tab:extend_ablation}
\setlength{\tabcolsep}{4pt}
\begin{tabular}{llcc}
\toprule
 \textbf{Configuration} & \textbf{Avg} & \textbf{$\Delta$} \\
\midrule
 \textbf{loss=[0,1,2,4]} & \textbf{0.5384} & --- \\
\midrule
 loss=[1,1,1,1]    & 0.5240 & $-0.0144$ \\
 loss=[1,1,2,4]    & 0.5247 & $-0.0137$ \\
 loss=[1,2,4,8]    & 0.5327 & $-0.0057$ \\
\bottomrule
\end{tabular}
\end{table*}

\begin{table*}[h!]
\centering
\caption{%
    \textbf{Biological-prior ablation under the same MSGR backbone} (9-dataset average).
    All variants share the identical model architecture and training protocol, differing only in the hierarchy construction strategy.
    $\Delta$ columns report the difference relative to MSGR-GO.
    ``PCC-200 Wins'' counts the number of datasets on which each variant achieves the highest PCC-200.
}
\label{tab:bio_prior_ablation}
\setlength{\tabcolsep}{4pt}
\begin{tabular}{lcccccccc}
\toprule
\textbf{Method} & \textbf{PCC-10} & \textbf{PCC-50} & \textbf{PCC-200} & \textbf{MSE}$\downarrow$ & \textbf{MAE}$\downarrow$ & $\Delta$PCC-200 & $\Delta$MSE & PCC-200 Wins \\
\midrule
\textbf{MSGR-GO}       & \textbf{0.776} & \textbf{0.710} & \textbf{0.517} & \textbf{1.348} & \textbf{0.865} & --      & --      & \textbf{6/9} \\
MSGR-Pathway           & 0.762          & 0.696          & 0.506          & 1.418          & 0.887          & -0.011  & +0.070  & 2/9 \\
MSGR-PPI               & 0.768          & 0.699          & 0.503          & 1.402          & 0.879          & -0.014  & +0.054  & 1/9 \\
MSGR-Coexp             & 0.764          & 0.695          & 0.500          & 1.394          & 0.880          & -0.017  & +0.046  & 0/9 \\
MSGR-Random            & 0.753          & 0.685          & 0.490          & 1.432          & 0.887          & -0.027  & +0.084  & 0/9 \\
\bottomrule
\end{tabular}
\end{table*}

\begin{table*}[h!]
\centering
\caption{Top 10 GO term--MSigDB Hallmark pathway correlations on SKCM, sorted by gene overlap.}
\label{tab:msigdb_interpretability}
\setlength{\tabcolsep}{5pt}
\begin{tabular}{llccl}
\toprule
\textbf{GO Term (MSGR)} & \textbf{MSigDB Hallmark} & \textbf{Overlap} & \textbf{Pearson $r$} & \shortstack{\textbf{Biological Implication in SKCM}} \\
\midrule
cellular process & EMT & 8 & 0.940 & Phenotype switching and metastasis \\
cellular process & KRAS Signaling Up & 7 & 0.923 & RAS/MAPK pathway activation \\
cellular process & G2/M Checkpoint & 5 & 0.924 & Cell-cycle progression \\
cellular anatomical structure & Apical Junction & 4 & 0.947 & Reduced adhesion and enhanced invasiveness \\
developmental process & EMT & 3 & 0.939 & Phenotype switching and metastasis \\
cellular anatomical structure & Mitotic Spindle & 3 & 0.925 & Active mitosis \\
protein-containing complex & G2/M Checkpoint & 3 & 0.922 & Cell-cycle progression \\
cellular anatomical structure & Hypoxia & 3 & 0.907 & Tumor hypoxia and angiogenesis \\
protein-containing complex & TNF$\alpha$ via NF-$\kappa$B & 3 & 0.895 & Pro-inflammatory and survival signaling \\
catalytic activity & Xenobiotic Metabolism & 3 & 0.855 & Metabolic adaptation and detoxification \\
\midrule
\multicolumn{5}{l}{\textbf{Mean Pearson $r$ of 10 shown pairs = 0.918}} \\
\multicolumn{5}{l}{\textbf{Mean Pearson $r$ of all 20 pairs = 0.896}} \\
\bottomrule
\end{tabular}
\end{table*}

\begin{table*}[h!]
\centering
\caption{Single-parent tree (MSGR) versus multi-parent averaging (MSGR-MP) on nine HEST-1k datasets (PCC-200).}
\label{tab:multiparent_projection}
\setlength{\tabcolsep}{3pt}
\begin{tabular}{lcccccccccc}
\toprule
& SKCM & HCC & LUNG & READ & COAD & IDC & PRAD & KIDNEY & CCRCC & \textbf{Avg} \\
\midrule
\textbf{MSGR} & \textbf{0.793} & \textbf{0.338} & \textbf{0.624} & \textbf{0.392} & \textbf{0.546} & \textbf{0.705} & \textbf{0.444} & \textbf{0.385} & \textbf{0.430} & \textbf{0.517} \\
MSGR-MP & 0.789 & 0.289 & 0.622 & 0.348 & 0.544 & 0.698 & 0.409 & 0.379 & 0.423 & 0.500 \\
\midrule
$\Delta$ & $-0.004$ & $-0.049$ & $-0.002$ & $-0.044$ & $-0.002$ & $-0.007$ & $-0.035$ & $-0.006$ & $-0.007$ & \textbf{$-0.017$} \\
\bottomrule
\end{tabular}
\end{table*}

\end{document}